\documentclass[10pt,journal,compsoc]{IEEEtran}
\ifCLASSOPTIONcompsoc
  \usepackage[nocompress]{cite}
\else
  \usepackage{cite}
\fi
\usepackage[dvipsnames]{xcolor}

\ifCLASSINFOpdf
  \usepackage[pdftex]{graphicx}
  \DeclareGraphicsExtensions{.pdf,.jpeg,.png,.svg}
\else
  \usepackage[dvips]{graphicx}
  \DeclareGraphicsExtensions{.eps,.pdf,.jpeg,.png,.svg}
\fi
\usepackage{algorithm}
\usepackage{algpseudocode}
\usepackage{multirow}
\usepackage{booktabs}
\usepackage{bbm}
\usepackage{bibentry}

\newtheorem{definition}{Definition}

\newtheorem{theorem}{Theorem}

\usepackage{amsmath}
\usepackage{amssymb}
\def\highlight{\textcolor{black}}
\newcommand\blfootnote[1]{%
  \begingroup
  \renewcommand\thefootnote{}\footnote{#1}%
  \addtocounter{footnote}{-1}%
  \endgroup
}

\begin{document}
%
\title{Semantic Invariant Multi-view Clustering with Fully Incomplete Information}

\author{Pengxin~Zeng,
        Mouxing~Yang,
        Yiding~Lu,
        Changqing~Zhang,
        Peng~Hu,
        and~Xi~Peng
}

%
%

\markboth{Journal of \LaTeX\ Class Files,~Vol.~14, No.~8, August~2015}%
{Shell \MakeLowercase{\textit{et al.}}: Bare Demo of IEEEtran.cls for Computer Society Journals}
%



\IEEEtitleabstractindextext{%
\begin{abstract} 
Robust multi-view learning with incomplete information has received significant attention due to issues such as incomplete correspondences and incomplete instances that commonly affect real-world multi-view applications. 
Existing approaches heavily rely on paired samples to realign or impute defective ones, but such preconditions cannot always be satisfied in practice due to the complexity of data collection and transmission. 
To address this problem, we present a novel framework called SeMantic Invariance LEarning (SMILE) for multi-view clustering with incomplete information that does not require any paired samples. 
To be specific, we discover the existence of invariant semantic distribution across different views, which enables SMILE to alleviate the cross-view discrepancy to learn consensus semantics without requiring any paired samples. 
The resulting consensus semantics remains unaffected by cross-view distribution shifts, making them useful for realigning/imputing defective instances and forming clusters. 
We demonstrate the effectiveness of SMILE through extensive comparison experiments with 13 state-of-the-art baselines on five benchmarks. 
Our approach improves the clustering accuracy of NoisyMNIST from 19.3\%/23.2\% to 82.7\%/69.0\% when the correspondences/instances are fully incomplete. The code could be accessed from \url{https://pengxi.me}.
\end{abstract}

}

\maketitle
\blfootnote{$^\dagger$ Pengxin Zeng, Mouxing Yang, Yiding Lu, Peng Hu, and Xi Peng are with College of Computer Science, Sichuan University, Chengdu, 610065, China. E-mail: \{zengpengxin.gm, yangmouxing, yidinglu.gm, penghu.ml, pengx.gm\}@gmail.com}

\blfootnote{$^\dagger$ Changqing Zhang is with College of Intelligence and Computing, Tianjin University, Tianjin, China. E-mail: zhangchangqing@tju.edu.cn}

\blfootnote{Corresponding author: Xi Peng.}

\IEEEdisplaynontitleabstractindextext

%
\IEEEpeerreviewmaketitle

\IEEEraisesectionheading{\section{Introduction}\label{Sec: introduction}}

%
%
%
%

\begin{figure}
\centering
\includegraphics[width=0.85\linewidth]{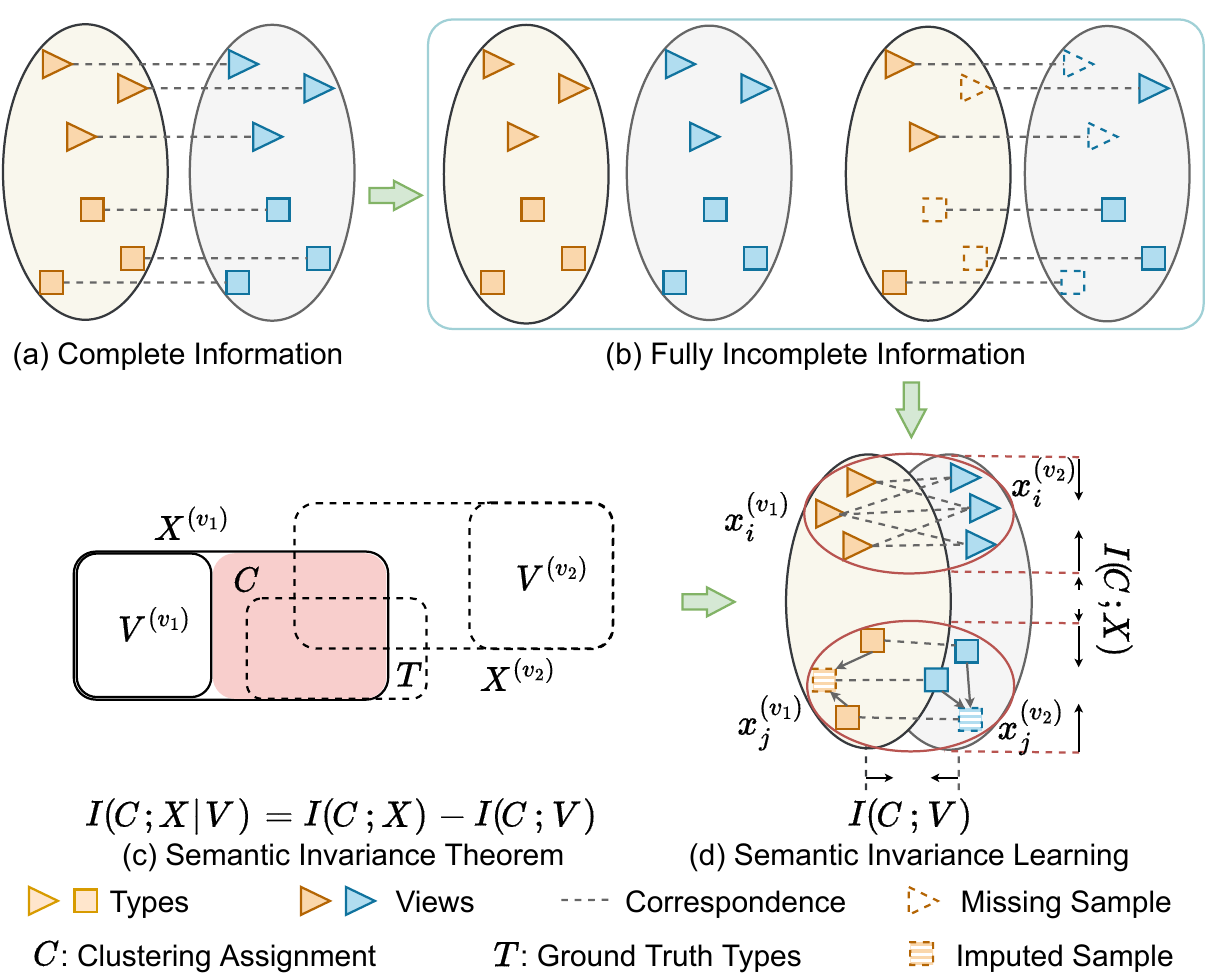}
 \vspace{-0.4cm}
\caption{Our motivation. Without loss of generality, we take two views as an example. In the figure, the dashed box indicates that the corresponding variable is unavailable or incomplete. \textbf{(a)} Complete information; \textbf{(b)} Fully incomplete information, \textit{i.e.}, either the correspondences or samples are missing for each instance; \highlight{\textbf{(c)} Information diagram of our Semantic Invariance Theorem. 
\textbf{(d)} Illustration on our Semantic Invariance Learning framework. In brief, it aims at maximizing $I(C; X|V) = I(C; X) - I(C; V)$ (the pink part) to simultaneously alleviate the cross-view discrepancy $I(C; V)$ and enhance the semantic discrimination $I(C; X)$. Thus, on the one hand, the incomplete correspondences could be rebuilt by associating cross-view samples with the same semantics. 
On the other hand, the missing samples could be imputed with the help of their semantic neighbours, which could be identified by the existing cross-view counterparts.
As a result, the defective instances could be realigned/imputed, and the cross-view clusters could be formed without requiring any paired samples.}}
\vspace{-0.7cm},
\label{Fig.1}
\end{figure}

\IEEEPARstart{M}{ulti-view} clustering (MvC)~\cite{wang2018detecting,xu2015multiselfpaced,kang2020large,lu2016convex,tao2017ensemble,wang2018partial} aims to alleviate the cross-view discrepancy while enhancing the semantic discrimination across different categories~\cite{vinokourov2002inferring,li2018survey}. 
Despite the rapid development of MvC, the successes of most MvC methods heavily rely on the assumption of complete information~\cite{zhang2019ae2,  peng2019comic,    wang2015deep, yin2020shared, yang2019split} (Fig.~\ref{Fig.1}(a)), \textit{i.e.}, the correspondences and instances are complete. In brief, the correspondences are complete if all samples are well aligned across views, and the instances are complete if all samples could be observed in all views. 
In practice, however, such an assumption is hard to satisfy due to the complexity of data collection and transmission. 

To address the aforementioned issue, various approaches have been proposed to explore how to learn from (partially) incomplete information. For incomplete correspondences, existing methods typically aim to re-align the unaligned samples using a permutation matrix~\cite{yu2021novel,gong2022gromov,huang2020partially} or their distance in hidden space~\cite{yang2021partially, karpathy2015deep, wei2020universal}. 
However, these methods built their success on an assumption of instance completeness, which is too ideal to satisfy in real scenarios. 
In contrast, some methods aim to learn a shared representation across all views without explicitly imputing the unobserved samples or representations~\cite{hu2019doubly, li2014partial, xu2015multi, shao2015multiple,liu2021efficient,zhang2020deep, xu2019adversarial}. To capture high nonlinearity, some approaches adopt deep neural networks to predict the representations of unobserved samples, embracing powerful learning and nonlinear modeling abilities~\cite{lin2022dual, lin2021completer, tang2022deep,yang2022robust}. Despite their promising performance, these methods still heavily rely on some well-aligned paired samples (\textit{i.e.}, both samples are observed and correspond correctly to each other), which are often unavailable in real-world applications. For example, when scouting a large area with several drones (views), the paired samples are almost impossible to obtain since each drone takes a separate reconnaissance route and the target is unlikely to exist in all views at the same time. Thus, it is still an open question to achieve multi-view clustering with \textbf{fully} incomplete information (Fig.~\ref{Fig.1}(b)).
 
In this paper, we propose a unified framework called SeMantic Invariance LEarning (SMILE), which is designed to achieve multi-view clustering in the presence of fully incomplete information. Specifically, our SMILE aims to alleviate the cross-view discrepancy while enhancing semantic discrimination, even in the absence of paired samples. To this end, we present the Semantic Invariance theorem (Theorem~\ref{Theorem.1}), namely \textit{the semantic distribution is invariant across different views}, which reveals the intrinsic property of multi-view clustering. This enables SMILE to alleviate the cross-view distribution discrepancy without requiring any paired samples, as each view takes supervision from the distributions of other views instead of certain cross-view pairs. Formally, SMILE formulates the cross-view discrepancy as $I(C; V)$ and the semantic discrimination as $I(C; X)$ as depicted in Fig.~\ref{Fig.1}(d). More specifically, $I(C; V)$ encourages the clustering assignments $C$ to be independent of the source-view variable $V$ and thereby alleviates the cross-view discrepancy. On the other hand, $I(C; X)$ maximizes the mutual information between the clustering assignments $C$ and the inputs $X$, thereby improving the semantic discrimination. Both of these terms do not require any paired samples and can be unified as $I(C; X|V) = I(C; X) - I(C; V)$ as depicted in Fig.\ref{Fig.1}(c), which enables SMILE to learn consensus semantics that is not confounded by cross-view distribution shifts. The learned consensus semantics can serve as a good ladder to realign/impute the defective instances and to form clusters, thus achieving multi-view clustering with fully incomplete information. Finally, we summarize the contributions and novelties of this work as follows.

\begin{itemize}
    \item To the best of our knowledge, we could be one of the first works to explore multi-view clustering with fully incomplete information. To address this issue, we propose a foundational theorem, Semantic Invariance, for robust multi-view learning, which enables us to take supervision from the distributions of other views without requiring paired samples.
    \item A novel Cross-view Semantic Invariance Learning framework is presented for multi-view clustering with incomplete information. We theoretically reveal that it could not only compensate for incomplete information but also facilitate MvC. (Theorem~\ref{Theorem.3}-~\ref{Theorem.4}).
    \item To verify the effectiveness of our method, we conducted extensive comparison experiments with $13$ competitive baselines on five datasets. In addition to comparisons on clustering quality, some experiments are conducted to quantitatively and visually investigate the proposed method by re-building/imputing the correspondences/samples.
\end{itemize}

\section{Related Works}
\label{Sec.2}
In this section, we briefly review some most related works on two topics: multi-view clustering and information theory. 

\subsection{Multi-view Clustering}
In recent years, there have been numerous studies on multi-view clustering, most of which implicitly or explicitly rely on the assumption of complete information. Based on this strong assumption, they can focus on extracting the shared semantics among the heterogeneous information across views in various ways~\cite{zhang2019ae2, zhang2018binary, yin2021cauchy, peng2019comic, andrew2013deep, zhou2019dual, vinokourov2002inferring, bach2002kernel, wang2015deep, yin2020shared, yang2019split, chen2020multi}. However, in practice, this assumption may be violated, resulting in the problem of incomplete information, which can be two-fold: incomplete correspondences and incomplete instances.

To learn with incomplete correspondences, many methods attempt to rebuild the cross-view correspondences with a permutation matrix. For example, Yu \textit{et al.}~\cite{yu2021novel} and Gong \textit{et al.}~\cite{gong2022gromov} assume that the graph structures should be consistent across views so that the permutation matrices could map the graph structure of one view to that of another view. In addition, Huang \textit{et al.}~\cite{huang2020partially} shuffle the aligned samples and then optimize the permutation matrix in a supervised fashion. Beyond the permutation matrices, some methods re-align the unaligned samples according to their distance in hidden space~\cite{yang2021partially,karpathy2015deep,wei2020universal}. However, all of the above methods rely on the assumption of instance completeness. As for the methods robust with incomplete instances, they can roughly be grouped into two mainstreams. For the first stream, they explore achieving multi-view clustering by learning a shared representation across all the views via non-negative matrix factorization (NMF)~\cite{hu2019doubly, li2014partial, xu2015multi, shao2015multiple}, multiple kernel k-means with incomplete kernels (MKKM-IK)~\cite{liu2021efficient}, adversarial learning~\cite{zhang2020deep, xu2019adversarial}, etc.
Meanwhile, the methods of the other stream embrace the powerful deep neural networks and thus predict the representations of the unobserved samples. For example, Jiang \textit{et al.}~\cite{jiang2019dm2c} learn the unobserved representations via adversarial learning. Lin \textit{et al.}~\cite{lin2022dual} train a projector in a supervised fashion to predict the unobserved representations. Tang \textit{et al.}~\cite{tang2022deep} and Yang \textit{et al.}~\cite{yang2022robust} fill the unobserved representations with the average of adjacent cross-view features. Although some promising results have been achieved by these studies, almost all of them still heavily rely on paired samples to learn the shared representation or to impute the unobserved representations. For example, Yang \textit{et al.}~\cite{yang2022robust} introduces noise-robust contrastive learning, which constructs positive/negative pairs from paired samples, resulting in other instances being abandoned during training. Besides, Jiang \textit{et al.}~\cite{jiang2019dm2c} study the problem of learning with fully incomplete instances but ignore the problem of (partially/fully) incomplete correspondences, which is a crucial part of the problem of incomplete information.  Although existing methods have achieved great success, to the best of our knowledge, this work could be one of the first studies to achieve multi-view clustering with fully incomplete information. 
\begin{figure*}
\centering
\includegraphics[width=0.85\linewidth]{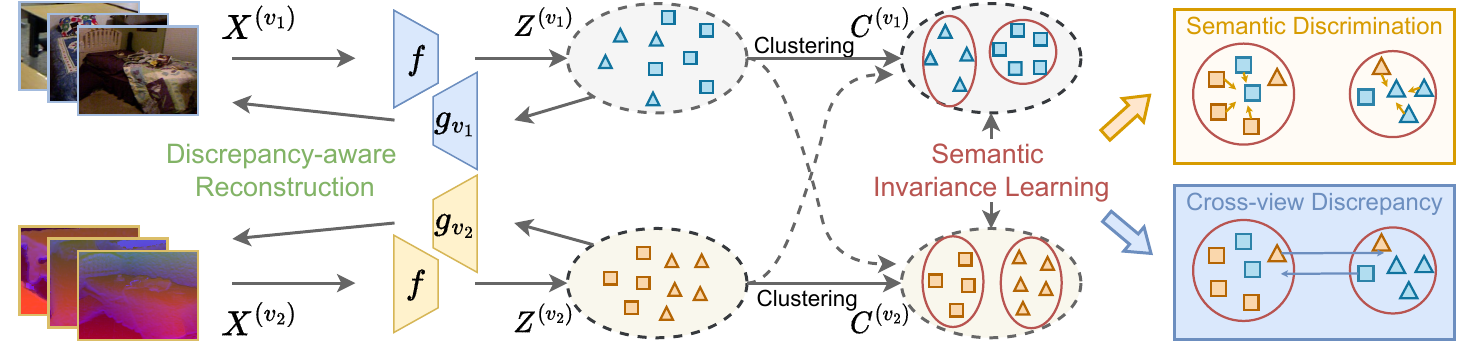}
\caption{The framework of our SMILE. \highlight{Without loss of generality, we take two views as an example.} SMILE intergrades two modules: the discrepancy-aware reconstruction module (DAR) and the semantic invariance learning module (SIL). DAR learns the view-specific representations by reconstructing the samples from their representations. SIL aims to alleviate the cross-view discrepancy while enhancing the semantic discrimination based on the clustering assignments on the view-specific representations.}
\label{Fig.2}
\end{figure*}
\subsection{Information Theory in Multi-view Learning}
Information-theory-based methods in multi-view learning have achieved promising achievements in recent years. These methods can be roughly classified into two streams. The first stream involves methods based on the information bottleneck~\cite{tishby2000information}, which enhance performance by explicitly or implicitly compressing learned representations to remove noisy information. For example, Wan \textit{et al.}~\cite{wan2021multi} and Federici \textit{et al.}~\cite{federici2020learning} compress the representation by explicitly minimizing $I(Z; X)$ and $I(Z^{(v_1)}; X^{(v_1)}|X^{(v_2)})$, respectively. In addition, Xu \textit{et al.}~\cite{xu2014large} present to compress the representation implicitly via a hidden layer with lower dimensionality than the last layer. The second stream of methods is based on contrastive learning~\cite{tian2020contrastive} which maximizes $I(Z^{(v_1)}, Z^{(v_2)})$ with various elaborate designs. For instance, Xu \textit{et al.}~\cite{xu2022multi} conduct contrastive learning separately in high-level feature space and label space to avoid conflict between learning consistent common semantics and reconstructing inconsistent view-private information. Additionally, Hassani \textit{et al.}~\cite{hassani2020contrastive} perform contrastive learning on multi-view graphs, contrasting the encodings from first-order neighbors and graph diffusion. Furthermore, Wang \textit{et al.}~\cite{wang2022rethinking} explore capturing more downstream task-relevant information by maximizing $I(Z; X)$ in addition to $I(Z^{(v_1)}|Z^{(v_2)})$. However, most of these methods focus on multi-view learning with complete information, which is hard to fully satisfy in real-world scenarios. To this end, Yang \textit{et al.}~\cite{yang2022robust} propose a robust contrastive term that identifies false negatives for MvC with partially incomplete information. 
Lin \textit{et al.}~\cite{lin2022dual,lin2021completer} induce a cross-view projector to learn data recovery and cross-view consistency as a whole. Although these methods have achieved promising results on the problem of partially incomplete information, they still heavily rely on paired samples. Different from the aforementioned methods, our method takes supervision from the variable $V$ to free our model from the assumption of information completeness entirely.

\section{Method}
In this section, we first present the formal definition of the fully incomplete information problem for multi-view clustering (MvC) in Sec.~\ref{Sec.3.1}.
In Sec.~\ref{Sec.3.2}, we elaborate on the cross-view semantic invariance theorem, which could not only compensate for incomplete information but also facilitate MvC with theoretical guarantees.
Based on the theorem in Sec.~\ref{Sec.3.3}, we propose a unified semantic invariance learning framework for MvC with fully incomplete information.

\subsection{Problem Formulation}
\label{Sec.3.1}
In this work, we explore how to achieve robust multi-view clustering with (fully) incomplete information, \textit{i.e.}, Partially Incomplete Information (PII), and Fully Incomplete Information (FII). We formulate the problem as follows:

\begin{definition} 
\textbf{Partially Incomplete Information.} 
A multi-view dataset $\{X^{(v)}\}^{M}_{v=1} = \{x^{(v)}_{1}, x^{(v)}_{2}, \dots, x^{(v)}_{N}\}^{M}_{v=1}$ consists of two subsets: i) $\{S^{(v)}\}^{M}_{v=1} = \{s^{(v)}_{1}, s^{(v)}_{2}, \dots, s^{(v)}_{N_s}\}^{M}_{v=1}$ with complete information, and ii) $\{W^{(v)}\}^{M}_{v=1} = \{w^{(v)}_{1}, w^{(v)}_{2}, \dots, w^{(v)}_{N_w}\}^{M}_{v=1}$ with either or both problems of incomplete correspondences and incomplete instances, where $N = N_s + N_w$ and $M$ denote the number of instances and views, respectively.
Specifically, the correspondences are incomplete if 
\begin{equation}
\sum_{v_1}^M \sum_{v_2 \neq v_1}^M Cor\left(\mathbf{w}_i^{\left(v_1\right)}, \mathbf{w}_i^{\left(v_2\right)}\right) < M(M-1), \forall i \in \left[1, N_w\right],
\end{equation}
where $Cor(a,b)$ is an indicator function evaluating to $1$ \textit{i.f.f.} samples $a$ and $b$ belong to the same instance. Besides, the instances are incomplete if  
\begin{equation}
1 \le |\{w^{(v)}_{i}\}_{v=1}^M| < M, \forall i \in \left[1, N_w\right],
\end{equation}
where $|\cdot |$ refers to the number of observed samples.
\end{definition}

\begin{definition} 
\textbf{Fully Incomplete Information (FII).}
A multi-view dataset $\{X^{(v)}\}^{M}_{v=1} = \{x^{(v)}_{1}, x^{(v)}_{2}, \dots, x^{(v)}_{N}\}^{M}_{v=1}$ with fully incomplete information only consists of $\{W^{(v)}\}^{M}_{v=1} = \{w^{(v)}_{1}, w^{(v)}_{2}, \dots, w^{(v)}_{N_w}\}^{M}_{v=1}$, where $N = N_w$. In other words, it is unavailable for paired samples \textit{i.e.}, both samples are observed and correspond correctly to each other.
\end{definition}
Although many approaches have been proposed to tackle the problem of partially incomplete information, existing approaches~\cite{yang2022robust, yang2021partially, lin2022dual, lin2021completer, tang2022deep, huang2020partially, liu2021efficient} still heavily rely on paired samples, which will hinder them to tackle real-world scenarios. In brief, it is still an open question to tackle the problem of fully incomplete information. 
In the following sections, we will elaborate on how to achieve the MvC with fully incomplete information. To be specific, we will first establish a cross-view semantic invariance theorem to shed light on the essence of fully incomplete information. Building upon this theorem, we then present a unified semantic invariance learning framework accordingly.

\subsection{Cross-view Semantic Invariance for MvC with FII}\label{Sec.3.2}
In this section, we start by proposing a foundational theorem, Cross-view Semantic Invariance, for robust multi-view learning in Sec.~\ref{Sec.3.2.1}. Based on the theorem, we theoretically reveal that the theorem could facilitate the solving of the problem of fully incomplete information in Sec.~\ref{Sec.3.2.2}. 
Finally, we theoretically reveal that the theorem could boost the clustering quality with theoretical guarantees by providing sufficient information for MvC in the meantime in Sec.~\ref{Sec.3.2.3}.

\subsubsection{Cross-view Semantic Invariance}
\label{Sec.3.2.1}
MvC aims to alleviate the cross-view discrepancy while enhancing semantic discrimination across different categories. However, for MvC with fully incomplete information, it is challenging to alleviate the cross-view discrepancy, since we cannot resort to paired samples to bridge the gap between different views. To address the challenge, we reveal that the distribution of ground-truth labels is independent of different views, which could be mathematically formulated as a foundational theorem (i.e., Theorem~\ref{Theorem.1}), termed Cross-view Semantic Invariance. 
\begin{theorem} \label{Theorem.1}
\textbf{Cross-view Semantic Invariance.} For multi-view data (with complete information or partially incomplete information or fully incomplete information), the distribution of the ground truth semantic category $T$ of the samples is invariant across different views $V$, \textit{i.e.}, mutual information $I(T(X); V) = 0$.
\end{theorem}
The proof of the theorem is provided in the appendix due to space limits. 
Theorem~\ref{Theorem.1} reveals that we could alleviate the cross-view discrepancy by enforcing the clustering assignments $C$ to be independent of the view $V$, \textit{i.e.}, minimizing $I(C; V)$. Notably, $I(C; V)$ just takes supervision from the distributions of the semantic categories in each view without any cross-view pairs, thus decoupling the dependence on paired samples to be immune against incomplete information. 
On the other hand, to enhance semantic discrimination, we replenish the statistical information shared between the clustering assignments $C$ and the input data $X$ via maximizing $I(C; X)$. Therefore, we could combine the two terms above as $I(C; X|V) = I(C; X) - I(C; V)$ to find the samples sharing the same semantics to compensate for incomplete information, \textit{i.e.}, re-building/imputing the correspondences/samples as proved in Theorem~\ref{Theorem.3} and \ref{Theorem.2}. Meanwhile, $I(C; X|V)$ can establish clustering-favorable clusters to boost the multi-view clustering quality, which is mathematically proved in Theorem~\ref{Theorem.4}. We dub $I(C; X|V)$ as cross-view Semantic Invariance Learning (SIL). 
In the following sections, we further theoretically prove that semantic invariance learning could not only tackle the fully incomplete information problem but also boost clustering quality.

\subsubsection{SIL Tackles the Fully Incomplete Information Problem} 
\label{Sec.3.2.2}
In this section, we theoretically prove that semantic invariance learning could facilitate solving the incomplete correspondences problem and the incomplete instances problem simultaneously. We present detailed proofs for both of these problems below:

\textbf{1. For the correspondence-incomplete data}, we formulate its solution to be a classification task, \textit{i.e.}, classifying $z_{i}^{(v_1)}$ into category $T(x_{i}^{(v_1)})$, where $z_{i}^{(v_1)}$ is the hidden representation of $x_{i}^{(v_1)}$. Since the essence of clustering is a one-to-many mapping and we could build correspondences between any samples belonging to the same category $T$~\cite{yang2022robust}. Based on the formulation, we consider the Bayes error rate $P_e$, which is the lowest achievable error for the given representations~\cite{fukunaga2013introduction}. Similar to the classification error rate in the representation learning~\cite{wang2022rethinking}, we deduce the Bayes error rate for solving incomplete correspondences as follows:
\begin{equation} \label{Eq.3.3.2.5}
    P_e = 1 - \mathbb{E}_{P{(z_{i}^{(v_1)})}} \max_{t\in T} P(T(z_{i}^{(v_1)}) = t).
\end{equation}
Based on this, we present the following theorem, which reveals the relationship between the cross-view semantic invariance and the incomplete correspondences:
\begin{theorem} \label{Theorem.3}
\textbf{Realigning the correspondence-incomplete data via Cross-view Semantic Invariance Learning.} Based on Theorem~\ref{Theorem.1}, the minimal achievable Bayes error rate $P_e$ for a given correspondence-incomplete dataset is bounded by the Semantic Invariance $I(C; X|V)$, \textit{i.e.}, 
\begin{equation}
    P_e \le  1 - \exp\left(-H(T, X|V) + I(C; X|V)\right),
\end{equation}
where $H(T, X|V)$ is a constant for a given dataset.
\end{theorem}
The theorem reveals that semantic invariance learning facilitates the resolution of the incomplete correspondence problem.

\textbf{2. For instance-incomplete data}, we formulate its solution to be a regression task, \textit{i.e.}, predicting the unobserved and continuous sample $x_{i}^{(v_2)}$ by using an observed feature of another view $z_{i}^{(v_1)}$. Based on this formulation, similar to the regression error in representation learning~\cite{wang2022rethinking}, we deduce the minimum achievable expected squared prediction error for solving incomplete instances as follows:
\begin{equation}\label{Eq.3.2.2.3}
\begin{aligned}
R_e = \min_{g_{v_2}} \mathbb{E}_{P{(z_{i}^{(v_1)})}}  ||x_{i}^{(v_2)} - g_{v_2}(z_{i}^{(v_1)})||^2,
\end{aligned}
\end{equation}
 where $g_{v_2}$, for simplicity, represents the mapping function between the features and the samples of view $v_2$ (refer to Lines $15$-$16$ of Algorithm~1 in Supplementary for more details). Based on this, we present the following theorem, which reveals the relationship between the cross-view semantic invariance and the problem of incomplete instances. 
\begin{theorem} \label{Theorem.2}
\textbf{Imputing instance-incomplete data via Cross-view Semantic Invariance Learning.} Based on Theorem~\ref{Theorem.1}, the lowest achievable expected squared prediction error $R_e$ for a given instance-incomplete dataset is bounded by the semantic invariance $I(C; X|V)$, \textit{i.e.}, 
\begin{equation} \label{Eq.3.2.2.4}
     R_e \le \alpha \cdot \exp\left(2H(T, X|V) - 2I(C; X|V)\right),
\end{equation}
where $H(T, X|V)$ is a constant for a given dataset, and $\alpha$ is also a constant.
\end{theorem}
This theorem reveals that semantic invariance learning facilitates the resolution of the incomplete instance problem.

In conclusion, we have provided theoretical proofs showcasing the ability of semantic invariance learning to simultaneously address the challenges of the incomplete correspondences problem and incomplete instances problems.

\subsubsection{SIL Boosts Clustering Quality}
\label{Sec.3.2.3}
Beyond addressing the problem of information incompleteness, we also theoretically prove that semantic invariance learning significantly enhances the quality of clustering by providing ample information for MvC. Specifically, we consider the lowest achievable clustering error rate, denoted as:
\begin{equation} \label{Eq.3.2.3.1}
    C_e = 1 - \sum_k      \max_{t \in T} |\tilde{T}_t \cap \tilde{C}_k| / |X|,
\end{equation}
where $\tilde{C}_k = \{x^{(v)}_i|C(x^{(v)}_i)=k\}$ denotes the set of samples assigned to $k$-th cluster, and $\tilde{T}_t = \{x^{(v)}_i|T(x^{(v)}_i)=t\}$ represents the set of samples belonging to $t$-th category. Building upon the aforementioned analysis, we present the following theorem, which reveals the relationship between semantic invariance learning and the clustering error rate: 
\begin{theorem}  \label{Theorem.4}
\textbf{Multi-view Clustering with Incomplete Information via Semantic Invariance Learning Learning.} Based on Theorem~\ref{Theorem.1}, the lowest achievable clustering error rate $C_e$ is bounded by the semantic invariance learning $I(C;X|V)$, \textit{i.e.}, 
\begin{equation}
    C_e \le  1 - \exp\left(-H(T, X|V)  + I(C;X|V)\right),
\end{equation}
where $H(T, X|V)$ is a constant for a given dataset, $T$ represents the ground-truth label variable, and $C$ denotes the clustering assignment variable (refer to in Supplementary Sec.~5 for details). 
\end{theorem}
The theorem demonstrates that maximizing semantic invariance learning $I(C; X|V)$ minimizes the lowest achievable clustering error rate $C_e$. When $I(C;X|V)$ is maximized (\textit{i.e.}, $I(C;X|V) = H(X|V)$), the information contained by $C$ becomes sufficient for MvC (\textit{i.e.}, $I(C;T) = I(X;T)$), leading to the achievement of minimal $C_e$. 

In summary, semantic invariance learning not only addresses the challenge of fully incomplete information but also enhances the clustering quality simultaneously, without necessitating any paired samples.

\subsection{Semantic Invariant MvC framework with FII}
\label{Sec.3.3}

Based on the theoretical analyses, we propose our unified semantic invariance learning framework SMILE for MvC with fully incomplete information in this section. As illustrated in Fig.~\ref{Fig.2}, SMILE integrates two modules: the discrepancy-aware reconstruction module (DAR) and the semantic invariance learning module (SIL). DAR reconstructs samples from their representations to learn view-specific representations, thus mitigating the dominance of cross-view discrepancy in the representations. Based on the view-specific representations, the clustering assignments are extracted for SIL, which alleviates the cross-view discrepancy while enhancing semantic discrimination. The overall loss function is summarized below:
\begin{equation} \label{Eq.3.3.1}
    \mathcal{L} = \lambda_{SIL}\mathcal{L}_{SIL} + \mathcal{L}_{DAR} ,
\end{equation}
where the $\lambda_{SIL}$ is a trade-off hyper-parameter fixed at $0.04$ for all datasets. In the following sections, we will elaborate on each loss item.

\subsubsection{Semantic Invariance Learning} \label{Sec.3.3.2}
The semantic invariance learning loss $\mathcal{L}_{SIL}$ aims to compensate for incomplete information and facilitates MvC simultaneously. To achieve this, we introduce a clustering assignment variable $C \in \mathbb{R}^{N \times M \times K}$, which models the likelihood of assigning $x_i^{(v)}$ to the $k$-th cluster. Based on this, our semantic invariance learning loss $\mathcal{L}_{SIL}$ could be formulated as follows:
\begin{equation} \label{Eq.3.3.2.2}
\begin{aligned}
    \mathcal{L}_{SIL} &= -I(C;X|V)\\
    &=- I(C; X) + I(C; V).
    \end{aligned}
\end{equation}
The first term $I(C; X)$ aims to enhance semantic discrimination across different categories. Specifically, let $\tilde{C}_k = \{x^{(v)}_i|C(x^{(v)}_i)=k\}$ denote the set of samples assigned to the $k$-th cluster, then we have
\begin{equation} 
\begin{gathered}
    \mathcal{L}_{SIL-s} = -I(C; X) = -H(C) + H(C|X) \\
    = \sum_k P{(\tilde{C}_k)}\log{P{(\tilde{C}_k)}} - \frac{1}{NM} \sum_{i,v,k}c^{(v)}_{ik}\log{c^{(v)}_{ik}},
    \end{gathered}
\end{equation}
where $P{(\tilde{C}_k)} =\frac{1}{NM} \sum_{i, v} c^{(v)}_{ik}$. Intuitively, minimizing $H(C|X)$ encourages the clusters to be \textbf{compact}, meaning that the intra-cluster distance should be smaller than the inter-cluster distance. However, this may lead to a trivial solution where all points are assigned to the same cluster. To avoid such a solution, we maximize $H(C)$ to encourage the clusters to be \textbf{balanced}, penalizing over-large or small clusters. By combining these two terms, $\mathcal{L}_{SIL-s}$ could enhance semantic discrimination across different categories.
  
The second term $I(C; X)$ is dedicated to alleviating the cross-view discrepancy. Specifically, let $\tilde{V}_v = \{x^{(j)}_i|j=v\}$ represent the set of samples belonging to the $v$-th view, then we have
  \begin{equation} 
   \begin{aligned}
    \mathcal{L}_{SIL-v} &= I(C;V) \\
    &= \sum_{k,v} P{(\tilde{C}_k, \tilde{V}_v)} \log{\frac{P{(\tilde{C}_k, \tilde{V}_v)}}{P{(\tilde{C}_k)}P{(\tilde{V}_v)}}},
     \end{aligned}
\end{equation}
where $P{(\tilde{V}_v)} = |\tilde{V}_v| / |X|$ and $P{(\tilde{C}_k, \tilde{V}_v)} = \frac{1}{N} \sum_{i} c^{(v)}_{ik}$. Minimizing $I(C; V)$ encourages the clusters to be \textbf{semantic-invariant}, meaning that the distribution of clustering assignments should be invariant across different views, thereby alleviating the cross-view discrepancy.



Based on the aforementioned analyses, we argue that $\mathcal{L}_{SIL-v}$ is a key component to directly alleviate the cross-view discrepancy. Therefore, we rewrite Equation~(\ref{Eq.3.3.2.2}) to explicitly highlight its role in our loss function, aiming to extract consensus semantics shared across views. The revised equation is as follows:
\begin{equation} \label{Eq.3.3.2.3}
    \mathcal{L}_{SIL} = \mathcal{L}_{SIL-s} + \gamma \mathcal{L}_{SIL-v},
\end{equation}
where $\gamma$ is a hyper-parameter that controls the balance between semantic discrimination and cross-view discrepancy alleviation in the learning process. 

\subsubsection{Discrepancy-Aware Reconstruction} \label{Sec.3.3.1}
In order to enhance the stability of semantic invariance learning, we present a reconstruction module to learn informative consensus representations $Z$ from the inputs $X$ and initialize the clustering assignments $C$ through k-means++ on $Z$. A vanilla implementation is to maximize $I(Z; X)$~\cite{autoencoder}, which could be formulated as:
\begin{equation} \label{Eq.4.4.1}
    \mathcal{L}_{Rec} = \mathbb{E}||x - \bar{g}(f(x))||^2,
\end{equation}
where $f$ and $\bar{g}$ denote the encoder and decoder, respectively. However, maximizing $I(Z; X)$ inevitably leads to an increase in the cross-view discrepancy at the representation level since $I(Z; X) = I(Z; X|V) + I(Z; V)$. To address the issue, we propose a novel discrepancy-aware reconstruction, which focuses on maximizing $I(Z; X|V)$ to learn informative consensus representation without introducing the cross-view discrepancy. The loss function could be formulated as follows:
\begin{equation} \label{Eq.3.3.1.1}
    \mathcal{L}_{DAR} = -I(Z;X|V) = -I(Z;X) + I(Z;V),
\end{equation}
where $-I(Z;X)$ and $I(Z;V)$ enhance semantic discrimination and alleviate the cross-view discrepancy at the feature level, respectively. Consequently, $I(Z; X|V)$ extracts consensus representations that are both discriminative and unaffected by the cross-view discrepancy. However, since $Z$ lies in a sparse space, directly optimizing $I(Z; V)$ is intractable. To overcome this, we rewrite it as:
\begin{equation}  \label{Eq.3.3.1.2}
    \mathcal{L}_{DAR} = -I(Z;X|V)= -H(X|V)+H(X|Z, V),
\end{equation}
where $H(X|V)$ is a constant term, and $H(X|Z, V) = -\mathbb{E}_{P_{(x, z, v)}} \log{P_{(x|z, v)}}$. Since approximating $P_{(x|z, v)}$ directly is intractable, we introduce a variational distribution $Q_{(x|z, v)}$ such that:
\begin{equation} \label{Eq.3.2.2}
\begin{aligned}
    H(X|Z, V) &= -\mathbb{E}_{P_{(x, z, v)}} \log{P_{(x|z, v)}} \\
    &= -\mathbb{E}_{P_{(x, z, v)}} \log{Q_{(x| z, v)}} \\
    &\quad - \mathbb{E}_{P_{(z, v)}}D_{KL} \left(P_{(x| z, v)} || Q_{(x| z, v)}\right) \\
    &\le -\mathbb{E}_{P_{(x, z, v)}} \log{Q_{(x| z, v)}},
\end{aligned}
\end{equation}
where $Q$ represents the variational distribution, which can be any type of distribution such as Gaussian~\cite{creswell2018generative} or Laplacian distribution~\cite{zhu2017unpaired}. For simplicity and considering the cross-view distribution discrepancy, we assume that the distribution $Q$ is a mixed Gaussian in our implementation. Specifically, we have: 
\begin{equation} \label{Eq.3.2.3}
    -\log Q_{(x| z, v)} \propto ||x - g_{v}(z)||^2,
\end{equation}
where $g_{v}$ maps a latent representation $z$ to the $v$-th Gaussian component corresponding to the distribution of the $v$-th view. By incorporating this formulation, we could rewrite Equation~(\ref{Eq.3.3.1.2}) as follows:
\begin{equation} \label{Eq.3.3.1.4}
    \mathcal{L}_{DAR} = \mathbb{E}||x - g(f(x))||^2,
\end{equation}
where $f(\cdot)$ denotes a shared encoder, and $g(\cdot) = g_{v}(\cdot)$ is a multi-branch decoder that handles the representations drawn from the $v$-th view. 

\begin{table*}
  \caption{Quantitative comparisons of SMILE with $13$ competitive baselines on five benchmarks under five settings. For each setting, the best and the second best results are marked in \textbf{bold} and \underline{underline}, respectively. \highlight{NS denotes the baselines that are not scalable to large datasets, and TvO represents the baselines that can only handle dual-view data.}}
  \label{Tab.1}
  \centering
  \setlength{\tabcolsep}{3.5pt}
\begin{tabular}{c|l|ccc|ccc|ccc|ccc|ccc}
\toprule
 \multirow{2}{*}{DataType} &\multirow{2}{4.5em}{Method} & \multicolumn{3}{c|}{NoisyMNIST} & \multicolumn{3}{c|}{MNISTUSPS} & \multicolumn{3}{c|}{Caltech} & \multicolumn{3}{c|}{CUB} & \multicolumn{3}{c}{\highlight{YouTubeFaces}} \\
 & & ACC & NMI & ARI & ACC & NMI & ARI & ACC & NMI & ARI & ACC & NMI & ARI & ACC & NMI & ARI \\ \midrule
 \multirow{2}{*}{\highlight{\shortstack{$100\%$ Unaligned\\($\zeta=100\%$)}}}  
 & MVC-UM~\cite{yu2021novel} & \underline{19.3} &\underline{ 9.9} &\underline{ 4.7} & \underline{53.5} & \underline{48.4} & \underline{35.0} & \underline{43.3} & \underline{67.3} & \underline{31.9} & \underline{44.3} & \underline{40.7} & \underline{23.0} & \highlight{NS} & \highlight{NS} & \highlight{NS} \\
 & GWMAC~\cite{gong2022gromov} & 11.4 & 0.3 & 0.1 &   15.6 &  3.7 &  1.5 &  4.9 &  16.0 &  0.3 &  28.3 &  21.0 &  9.1 & \highlight{\underline{3.2 }} & \highlight{\underline{2.3}} & \highlight{\underline{0.2}}\\
 & SMILE & \textbf{82.7} & \textbf{79.5} & \textbf{74.2} & \textbf{85.2} & \textbf{80.8} & \textbf{76.1} & \textbf{47.6} & \textbf{74.0} & \textbf{33.0} & \textbf{63.4} & \textbf{61.9} & \textbf{48.2} & \highlight{\textbf{52.5}} & \highlight{\textbf{73.6}} & \highlight{\textbf{42.6}} \\ \midrule
 \multirow{2}{*}{\highlight{\shortstack{$100\%$ Missing\\($\eta=100\%$)}}}  
 & DM2C~\cite{jiang2019dm2c} & \underline{23.2} & \underline{15.4} &\underline{ 8.0} & \underline{35.1} & \underline{34.2} & \underline{18.3} & \underline{28.2} & \underline{59.3} & \underline{18.3} & \underline{35.6} & \underline{36.4} &\underline{ 6.4} &\highlight{\underline{16.2}} & \highlight{\underline{32.1}} & \highlight{\underline{5.8}} \\
 & SMILE & \textbf{69.0} & \textbf{63.8} & \textbf{54.1} & \textbf{74.3} & \textbf{69.6} & \textbf{61.8} & \textbf{30.5} & \textbf{60.1} & \textbf{20.4} & \textbf{40.2} & \textbf{37.5} & \textbf{20.8} & \highlight{\textbf{26.5}}  & \highlight{\textbf{49.9}} & \highlight{\textbf{18.5}} \\ \midrule
 \multirow{12}{*}{\highlight{\shortstack{$50\%$ Unaligned\\($\zeta=50\%$)}}}  
 & DCCAE~\cite{wang2015deep} & 27.6 & 19.5 & 10.0 & 66.1 & 52.1 & 47.4 & 26.6 & 50.1 & 25.2 & 15.8 & 2.8 & 0.2 & \highlight{18.0} & \highlight{24.3} & \highlight{7.0} \\
 & BMVC~\cite{zhang2018binary} & 28.5 & 24.7 & 14.2 & 36.9 & 15.9 & 12.1 & 29.1 & 34.8 & 12.9 & 16.0 & 3.4 & 0.2 & \highlight{24.0} & \highlight{19.2} & \highlight{8.4} \\
 & AE2-Nets~\cite{zhang2019ae2} & 38.3 & 34.3 & 22.0 & 37.6 & 23.9 & 16.1 & 4.2 & 13.5 & 0.0 & 14.5 & 2.6 & 0.3 & \highlight{18.3} & \highlight{15.8} & \highlight{6.4} \\
 & DAIMC~\cite{hu2019doubly} & 37.6 & 34.3 & 22.8 & 44.3 & 34.5 & 24.8 & \underline{48.5} & 68.7 & 33.1 & 15.7 & 2.8 & 0.0 & \highlight{NS} & \highlight{NS} & \highlight{NS} \\
 & EERIMVC~\cite{liu2021efficient} & 46.8 & 29.6 & 23.9 & 53.3 & 37.4 & 31.9 & 26.4 & 36.5 & 9.2 & 15.8 & 2.9 & 0.0 & \highlight{NS} & \highlight{NS} & \highlight{NS} \\
 & PMVC~\cite{li2014partial} & 31.9 & 21.4 & 13.0 & 54.5 & 44.4 & 35.9 & 45.0 & 68.6 & 32.4 & 15.8 & 3.0 & 0.0 & \highlight{TvO} & \highlight{TvO} & \highlight{TvO} \\
 & PVC~\cite{huang2020partially} & 81.8 & 82.3 & 82.0 & 86.5 & 78.1 & 74.6 & 18.6 & 48.9 & 14.6 & 50.2 & 56.3 & 38.6 & \highlight{NS} & \highlight{NS} & \highlight{NS} \\
 & MvCLN~\cite{yang2021partially} & 91.1 & 84.2 & 83.6 & 90.0 & 81.4 & 80.4 & 35.6 & 61.0 & \textbf{40.9} & 58.2 & 55.2 & 40.8 & \highlight{54.0} & \highlight{\underline{69.2}} & \highlight{\underline{44.2}} \\
 & SURE~\cite{yang2022robust} & \underline{95.2} & \underline{88.2} & \underline{89.7} & \underline{92.1} & \underline{82.8} & \underline{83.5} & 46.2 & \underline{70.7} & {33.0} & \underline{64.5} & \underline{62.0} & \underline{47.9} & \highlight{\underline{54.7}} & \highlight{68.8} & \highlight{43.4} \\
 & DCP~\cite{lin2022dual} & 32.3 & 28.0 & 9.4 & 41.4 & 34.0 & 13.4 & 22.2 & 48.6 & 19.2 & 35.4 & 30.7 & 8.1 & \highlight{26.4} & \highlight{54.2} & \highlight{19.2} \\
 & DSIMVC~\cite{tang2022deep} & 34.6 & 24.0 & 16.8 & 62.2 & 47.4 & 39.7 & 20.6 & 31.0 & 16.3 & 30.4 & 25.4 & 11.8 & \highlight{21.0} & \highlight{33.8} & \highlight{10.9} \\
 & GWMAC~\cite{gong2022gromov} & 11.4 & 0.2 & 0.1 &  16.1 &  4.0 &  1.8 &  4.4 &  15.4 &  0.4 &  30.6 &  27.2 &  12.2 &  \highlight{3.2}
 &\highlight{ 2.2} & \highlight{0.2}\\
 & SMILE & \textbf{97.9} & \textbf{94.2} & \textbf{95.4} & \textbf{98.6} & \textbf{96.3} & \textbf{97.0} & \textbf{50.9} & \textbf{79.4} & \underline{35.2} & \textbf{71.1} & \textbf{70.4} & \textbf{58.2} & \textbf{\highlight{57.8}} & \textbf{\highlight{77.1}} & \textbf{\highlight{48.8}} \\ \midrule
 \multirow{12}{*}{\highlight{\shortstack{$50\%$ Missing\\($\eta=50\%$)}}}  
 & DCCAE~\cite{wang2015deep} & 65.4 & 62.9 & 38.3 & 79.5 & 79.2 & 68.4 & 29.1 & 58.8 & 23.4 & 42.3 & 40.9 & 25.5 & \highlight{19.0} & \highlight{37.9} & \highlight{8.6} \\
 & BMVC~\cite{zhang2018binary} & 30.7 & 19.2 & 10.6 & 43.9 & 39.0 & 21.0 & 40.0 & 58.5 & 10.2 & 29.8 & 20.3 & 6.4 & \highlight{34.1} & \highlight{42.7} & \highlight{7.4} \\
 & AE2-Nets~\cite{zhang2019ae2} & 29.9 & 23.8 & 11.8 & 40.9 & 29.3 & 19.7 & 6.6 & 18.0 & 4.5 & 35.9 & 32.0 & 15.9 & \highlight{18.8} & \highlight{27.9} & \highlight{8.5} \\
 & DAIMC~\cite{hu2019doubly} & 33.8 & 26.4 & 16.0 & 55.2 & 49.6 & 38.6 & \textbf{56.2}  & \underline{78.0}     & \underline{41.8} & 62.7 & 58.5 & 47.7 & \highlight{NS} & \highlight{NS} & \highlight{NS} \\
 & EERIMVC~\cite{liu2021efficient} & 55.6 & 45.9 & 36.8 & 65.2 & 55.7 & 48.9 & 43.6 & 69.0 & 26.4 & \underline{68.7} & \underline{63.9} & \underline{53.8} & \highlight{NS} & \highlight{NS} & \highlight{NS} \\
 & PMVC~\cite{li2014partial} & 33.1 & 25.5 & 14.6 & 60.5 & 47.1 & 39.8 & {48.4} & {72.8} & {40.4} & 57.7 & 54.4 & 38.3 & \highlight{TvO} & \highlight{TvO} & \highlight{TvO} \\
 & PVC~\cite{huang2020partially} & 16.4 & 6.7 & 2.3 & 14.7 & 4.4 & 1.4 & 6.6 & 17.4 & 0.3 & 39.0 & 40.5 & 20.9 & \highlight{NS} & \highlight{NS} & \highlight{NS} \\
 & MvCLN~\cite{yang2021partially} & 53.8 & 50.6 & 28.5 & 46.8 & 44.6 & 21.8 & 27.2 & 47.5 & 23.5 & 45.2 & 40.8 & 21.9 & \highlight{36.1} & \highlight{\underline{48.2}} & \highlight{23.7} \\
 & SURE~\cite{yang2022robust} & \underline{93.0} & \underline{85.4} & \underline{85.9} & {92.3} & {85.0} & {84.3} & {34.6} & {57.8} & {19.9} & {58.3} & {50.4} & {37.4} & \highlight{\underline{45.2}} & \highlight{46.9 }& \highlight{\underline{29.6}} \\
 & DCP~\cite{lin2022dual} & 80.0 & 75.2 & 70.7 & 94.0 & 89.7 & 88.3 & 44.3 & {71.0} &\textbf{ 45.3} & 53.7 & 65.5 & 47.3 & \highlight{26.3} & \highlight{47.2} & \highlight{14.4} \\
 & DSIMVC~\cite{tang2022deep} & 55.8 & 55.1 & 43.0 & \underline{97.0} & \underline{92.4} & \underline{93.5} & 16.4 & 24.8 & 9.2 & 54.4 & 52.4 & 35.2 & \highlight{29.4} & \highlight{48.5}  & \highlight{19.0} \\
 & SMILE & \textbf{96.8} & \textbf{91.7} & \textbf{93.0} & \textbf{98.5} & \textbf{95.7} & \textbf{96.6} & \underline{51.2} & \textbf{79.0} & {35.6} & \textbf{69.5} & \textbf{66.7} & \textbf{54.9} & \highlight{\textbf{54.6}} & \highlight{\textbf{76.3}} & \highlight{\textbf{45.2}} \\ \midrule
 \multirow{12}{*}{\shortstack{Complete\\Information}} 
 & DCCAE~\cite{wang2015deep} & 78.0 & 81.2 & 68.2 & 96.8 & 97.7 & 96.6 & 45.8 & 68.6 & {37.7} & 55.3 & 58.7 & 45.1 & \highlight{32.2} & \highlight{61.5} & \highlight{19.0} \\
 & BMVC~\cite{zhang2018binary} & 88.3 & 77.0 & 76.6 & 87.1 & 84.5 & 82.0 & 50.1 & 72.4 & 33.9 & 66.2 & 61.7 & 48.7 & \highlight{48.5} & \highlight{62.4} & \highlight{36.1} \\
 & AE2-Nets~\cite{zhang2019ae2} & 42.1 & 43.4 & 30.4 & 54.0 & 46.5 & 35.4 & 4.0 & 13.6 & 0.0 & 48.8 & 46.7 & 30.5 & \highlight{21.8} & \highlight{34.0} & \highlight{12.2} \\
 & DAIMC~\cite{hu2019doubly} & 38.4 & 34.7 & 23.0 & 65.1 & 65.5 & 54.2 & \textbf{57.5} & \underline{78.7} & \underline{41.9} & 71.6 & 70.7 & 57.9 & \highlight{NS} & \highlight{NS} & \highlight{NS} \\
 & EERIMVC~\cite{liu2021efficient} & 65.7 & 57.6 & 51.3 & 79.0 & 68.1 & 62.4 & 49.0 & 74.2 & 34.2 & \underline{74.0} & \underline{73.1} & \underline{62.4} & \highlight{NS} & \highlight{NS} & \highlight{NS} \\
 & PMVC~\cite{li2014partial} & 41.1 & 36.4 & 24.5 & 60.4 & 59.5 & 47.3 & 49.4 & 73.5 & 39.7 & 64.5 & 70.3 & 53.1 & \highlight{TvO} & \highlight{TvO} & \highlight{TvO} \\
 & PVC~\cite{huang2020partially} & 87.1 & 92.8 & 93.1 & 95.3 & 90.4 & 90.1 & 20.5 & 51.4 & 15.7 & 59.7 & 65.3 & 51.6 & \highlight{NS} & \highlight{NS} & \highlight{NS} \\
 & MvCLN~\cite{yang2021partially} & 97.3 & 94.2 & 95.3 & 98.8 & 96.5 & 97.3 & 39.6 & 65.3 & 32.8 & 59.7 & 56.5 & 42.5 & \highlight{\underline{57.3}} & \highlight{70.9} & \highlight{\underline{48.2}} \\
 & SURE~\cite{yang2022robust} & \underline{98.4} & \underline{95.4} & \underline{96.5} & \underline{99.1} & \underline{97.5} & \underline{98.1} & {43.8} & {70.1} & {29.5} & {58.0} & {59.3} & {45.2} & \highlight{55.6} & \highlight{\underline{75.8}} & \highlight{46.8} \\
 & DCP~\cite{lin2022dual} & 89.1 & 88.9 & 85.5 & 94.8 & 93.9 & 90.5 & \underline{51.3} & 74.8 & \textbf{51.9} & 63.6 & 70.2 & 53.9 & \highlight{34.0} & \highlight{60.2} & \highlight{16.5} \\
 & DSIMVC~\cite{tang2022deep} & 61.0 & 58.1 & 46.7 & 98.5 & 96.7 & 96.7 & 19.7 & 40.0 & 19.7 & 58.5 & 56.3 & 39.9 & \highlight{22.2} & \highlight{38.0} & \highlight{13.1} \\
 & GWMAC~\cite{gong2022gromov} & 11.3 & 0.3 & 0.1 &  14.5 &  2.8 &  1.1 &  4.4 &  15.1 &  0.2  &  29.1 &  21.8 &  10.0 & \highlight{3.1}  & \highlight{2.1} & \highlight{0.2} \\
 & SMILE & \textbf{99.3} & \textbf{97.8} & \textbf{98.4} & \textbf{99.9} & \textbf{99.7} & \textbf{99.8} & {51.0} & \textbf{79.4} & {35.3} & \textbf{74.7} & \textbf{75.5} & \textbf{64.5} & \highlight{\textbf{58.5}}& \highlight{\textbf{78.4}}& \highlight{\textbf{50.2}} \\
 \bottomrule
 \end{tabular}
\end{table*}
\section{Experiments}
In this section, we evaluate the effectiveness of our SMILE against the problem of (fully) incomplete information compared with $13$ state-of-the-art multi-view clustering methods on five benchmarks. In the following sections, we will elaborate on our experimental setting in Sec.~\ref{Sec.4.1}. 
Then, we will quantitatively verify the effectiveness of the proposed SMILE in Sec.~\ref{Sec.4.2}. 
Beyond the quantitative comparisons on clustering quality, more in-depth explorations will be conducted in Sec.~\ref{Sec.4.3}. Finally, we will conduct the ablation studies in Sec.~\ref{Sec.4.4} to shed some light on the essence of our SMILE. 
\begin{figure*}
\centering
\includegraphics[width=0.85\textwidth]{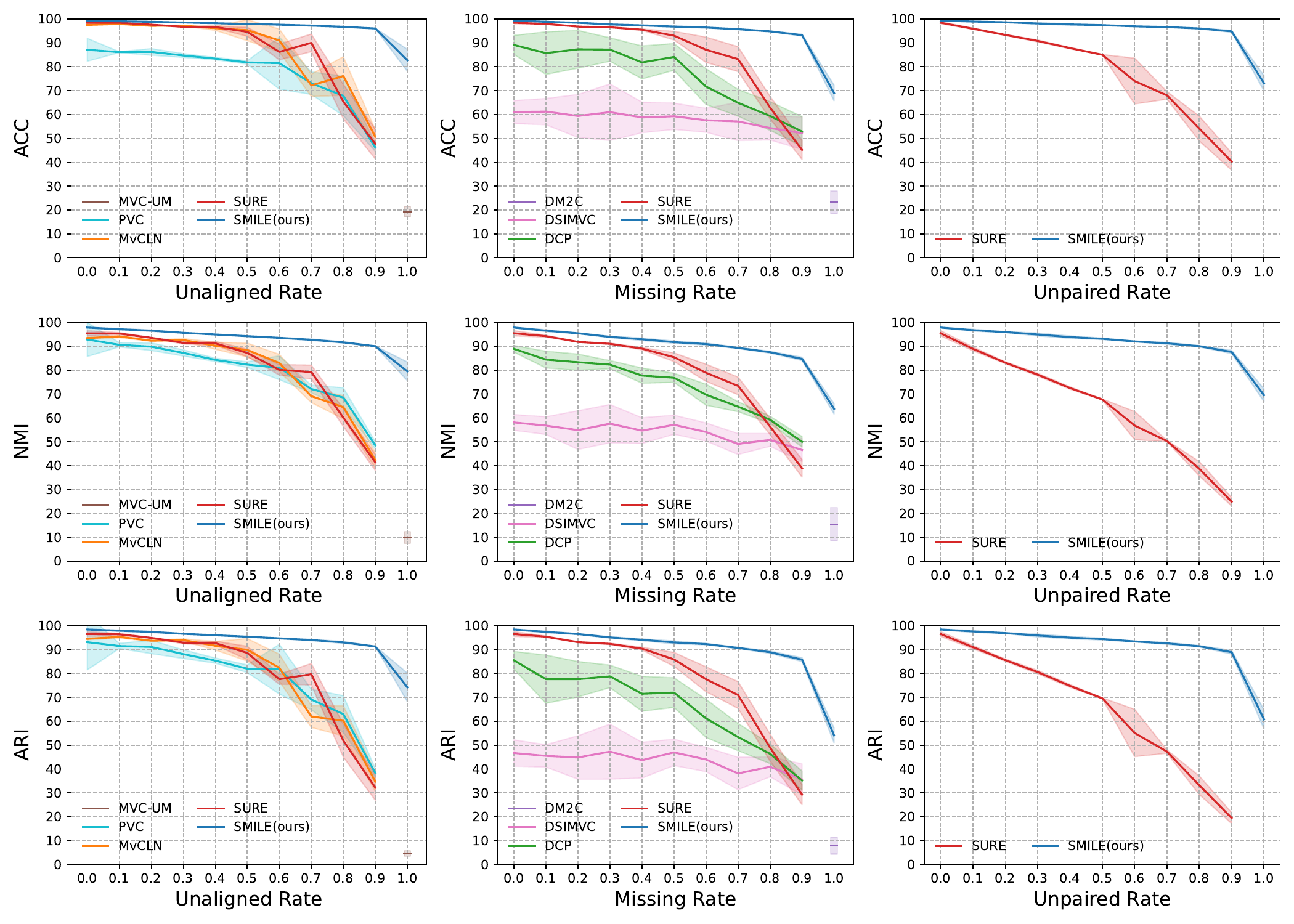}
\vspace{-0.3cm}
\caption{\highlight{Performance analysis on NoisyMNIST with different unaligned rates, missing rates, and unpaired rates.}}
\vspace{-0.3cm}
\label{Fig.3}
\end{figure*}

\subsection{Experimental Setups}
\label{Sec.4.1}

\textbf{Implementation Details:}  In our implementation, we set  $\lambda_{SIL} = 0.04$, $\gamma = 5$ in Equation~(\ref{Eq.3.3.2.3}). 
Moreover, we use a convolutional auto-encoder for multi-view image datasets, \textit{i.e.}, MNISTUSPS and NoisyMNIST, and a fully connected auto-encoder for other datasets. For each auto-encoder that contains a shared encoder, we add an additional adaption layer to accommodate the different input dimensions of each view. All the networks use the Adam optimizer with an initial learning rate of $1e-3$ for all datasets under all settings. In addition to handling fully incomplete information, we also conduct experiments under different settings where paired samples are provided for comprehensive comparisons. In those experiments, we incorporate contrastive learning into our method for fair comparisons. Finally, all the quantitative results of our SMILE are the average of five random seeds by default.

\textbf{Dataset:} We evaluate our method on five datasets, which are as follows:
\begin{itemize}
\item \textbf{NoisyMNIST~\cite{wang2015deep}:} This dataset contains $70,000$ instances, where each instance consists of two views: the raw MNIST image and its rotated and Gaussian noised version. For a fair comparison, we follow the previous work SURE~\cite{yang2022robust}, and randomly select $30,000$ instances for evaluation since some baselines cannot deal with such a large-scale dataset.
\item \textbf{MNISTUSPS:} This dataset includes $67,291$ images of digits from the MNIST and USPS datasets. Following~\cite{peng2019comic}, we randomly select $5,000$ samples from each dataset, distributed over $10$ digits.
\item \textbf{Deep Caltech-101:} This dataset consists of $8,677$ images of objects belonging to $101$ classes, with $100$ classes for objects and one class for background cluster. Following~\cite{han2021trusted}, we utilize deep features extracted by DECAF~\cite{krizhevsky2017imagenet} and VGG19~\cite{simonyan2014very} networks as two views.
\item \textbf{CUB~\cite{wah2011caltech}:} This dataset comprises various categories of birds. Following~\cite{zhang2020deep}, we employ deep visual features extracted by GoogLeNet and text features extracted by doc2vec~\cite{le2014distributed} as two views.
\item \highlight{\textbf{YouTubeFaces}~\cite{wolf2011face}: This dataset contains $152,549$ faces from $66$  identities, \textit{i.e.}, each people has more than $1,500$ face images at least. For comparisons, we describe each image using multi-view features consisting of $512$-dim GIST feature, $1984$-dim HOG feature, and $1024$-dim HIST feature. }
\end{itemize}

\textbf{Baselines:} We compare SMILE with $13$ competitive multi-view clustering baselines. Specifically, DCCAE~\cite{wang2015deep}, BMVC~\cite{zhang2018binary}, and AE2-Nets~\cite{zhang2019ae2} are designed for multi-view clustering with complete information. 
PVC~\cite{huang2020partially} and MvCLN~\cite{yang2021partially} are designed for partial correspondence incompleteness. Five baselines are designed for partial instance incompleteness, including PMVC~\cite{li2014partial}, DAIMC~\cite{hu2019doubly}, EERIMVC~\cite{liu2021efficient}, DCP~\cite{lin2022dual}, and DSIMVC~\cite{tang2022deep}. SURE~\cite{yang2022robust} is designed against partial information incompleteness. MVC-UM~\cite{yu2021novel} and DM2C~\cite{jiang2019dm2c} are designed against full correspondence incompleteness and full instance incompleteness, respectively. Since many baselines cannot handle partial correspondence/instance incompleteness directly, we follow SURE~\cite{yang2022robust} and adopt the following two approaches for fair comparisons: 
\begin{itemize}
\item For the baselines that cannot handle the partial correspondence incompleteness, we re-align the unaligned samples via the Hungarian algorithm~\cite{kuhn1955hungarian}. More specifically, we first obtain the PCA features of the samples and then use the Hungarian algorithm with the  Euclidean similarity to establish correspondences.
\item For the baselines that cannot handle the partial instance incompleteness, we fill the unobserved samples from the $v$-th view with the average of all the exiting samples of the view. 
\end{itemize}

\begin{figure*}
\centering
\includegraphics[width=0.85\textwidth]{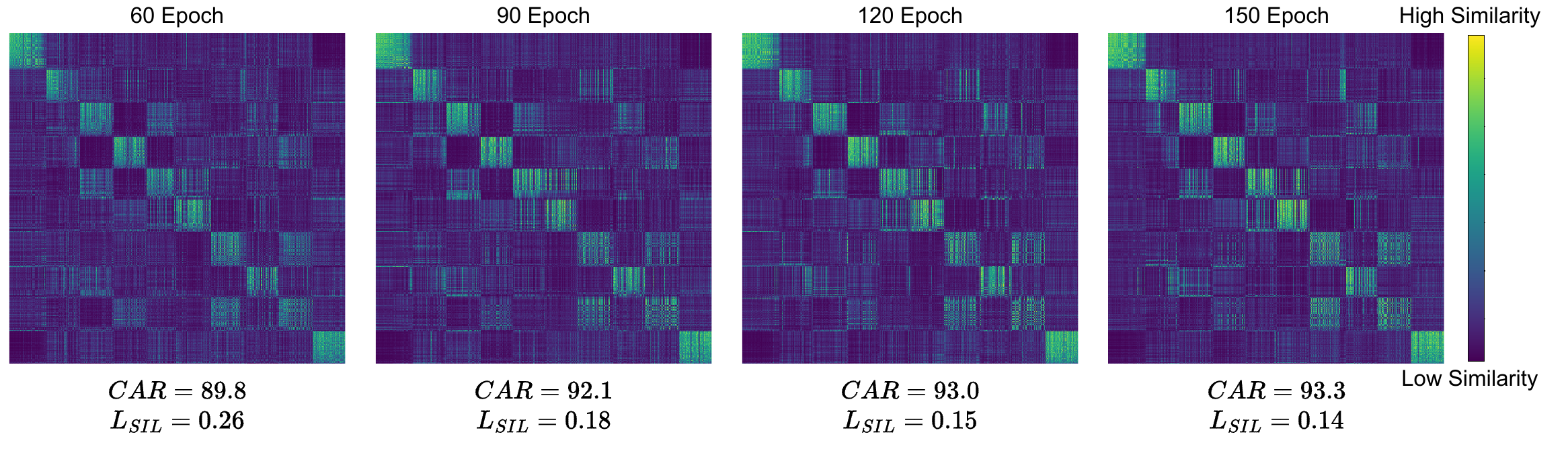}
\vspace{-0.3cm}
\caption{\highlight{The visualization of the similarity matrix on NoisyMNIST with the unaligned rate of $100\%$.} The similarity score in the $i$-th row $j$-th column denotes the similarity between the $i$-th unaligned sample in view $1$ and the $j$-th unaligned sample in view $2$. In each view, samples are sorted according to their categories.}
\label{Fig.4}
\vspace{-0.3cm}
\end{figure*}

\subsection{Quantitative Comparisons}
\label{Sec.4.2}

In this section, we conduct quantitative experiments to compare our SMILE with $13$ baselines under various missing rates and unaligned rates. To be specific, the missing rate is defined as $\eta = \frac{m}{N}$, where $N$ is the size of the dataset and $m$ is the number of instances with missing samples. \highlight{To generate the data with missing samples, we randomly choose $m$ instances and discard one sample/view for the instances, following the setting used in SURE~\cite{yang2022robust}.} Regarding the unaligned rate, it is defined as $\zeta = \frac{c}{N}$, where $c$ is the number of instances with incorrect correspondences. \highlight{To generate the data with incorrect correspondences, we also follow SURE~\cite{yang2022robust} to randomly sample $c$ instances and remove the correspondences between their samples.}

\highlight{We present the quantitative results in Table~\ref{Tab.1} (see Supplementary Material, Sec.~7 for more results)}. As shown in the table, there are two previous works, to our best knowledge, that could achieve multi-view clustering with the unaligned rate of $100\%$  --- MVC-UM~\cite{yu2021novel} and GWMAC~\cite{gong2022gromov}. Our SMILE outperforms them by a large margin by taking advantage of deep neural networks. With the missing rate of $100\%$, although DM2C~\cite{jiang2019dm2c} also incorporates deep learning, it is outperformed by our SMILE by a large margin on all five datasets. We conjecture that the superior performance of our SMILE is due to the utilization of information theory-based optimization instead of adversarial learning, thus being less prone to degenerate. With the missing rate of $50\%$ and unaligned rate of $50\%$, our SMILE outperforms the most competitive baselines on all the datasets in terms of ACC and NMI. We attribute this to the fact that we utilize unpaired samples (due to incomplete correspondences or incomplete instances) for training, while many competitive baselines brutally discard them~\cite{yang2022robust, lin2022dual,yang2021partially, huang2020partially}. In the setting of complete information, our SMILE also outperforms almost all baselines in the five datasets. The superiority of our method could be attributed to the unified and effective information theory-based framework. Overall, our SMILE achieves state-of-the-art performance in almost all settings. 

To further evaluate the effectiveness and robustness of our SMILE against incomplete information, we conduct performance analyses comparing SMILE with the most competitive methods in Fig.~\ref{Fig.3} under various unaligned/missing/unpaired rates. The unpaired rate $\varrho$ refers to the scenario where both correspondences and instances are incomplete, \textit{i.e.}, $\eta = \zeta = \varrho/2$. From the figure, it can be observed that all baselines heavily rely on paired samples, and their performances drop severely as the unaligned/missing/unpaired rate increases, reaching an accuracy of approximately $50\%$ when the rate reaches $90\%$. However, our SMILE maintains its performance with accuracy consistently above $93\%$ under the same settings. This can be attributed to our utilization of unpaired samples (due to incomplete correspondences or incomplete instances) for training, while most baselines brutally discard them, \textit{e.g.}, PVC~\cite{huang2020partially}, MvCLN~\cite{yang2021partially}, SURE~\cite{yang2022robust}, and DCP~\cite{lin2022dual}. Specifically, both $\mathcal{L}_{DAR}$ and $\mathcal{L}_{SIL}$ are calculated using all samples, even if some of them are unpaired. Therefore, our SMILE exhibits great effectiveness against incomplete information. Moreover, the standard deviation of our method is smaller than that of most baselines, demonstrating the robustness of our method. We conjecture that the semantic invariance learning loss $\mathcal{L}_{SIL}$ alleviates the randomness introduced by k-means by encouraging the learned representation clusters to be well-balanced, compact, semantically invariant, as analyzed in Sec.~\ref{Sec.3.3.2}. 

\begin{figure}
\centering
\includegraphics[width=0.85\linewidth]{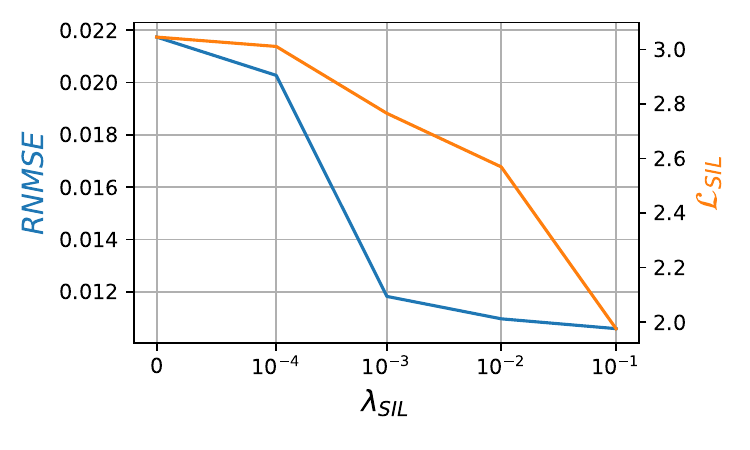}
\vspace{-0.5cm}
\caption{\highlight{Imputation performance analysis with the missing rate of $100\%$.} }
\label{Fig.5}

\end{figure}
\begin{figure}
\centering
\includegraphics[width=0.85\linewidth]{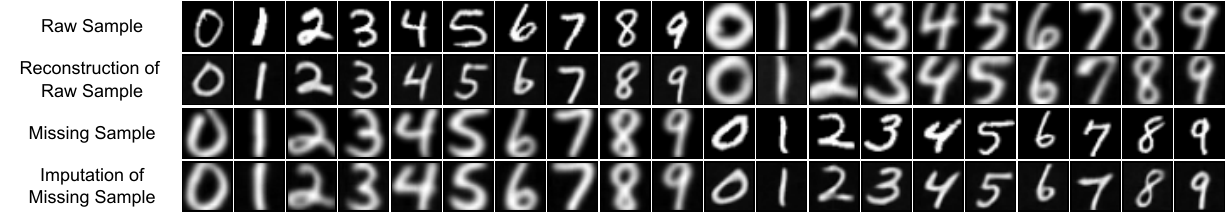}
\caption{\highlight{The imputation of the unobserved samples with the missing rate of $100\%$}. The raw samples are the inputs $x_i^{(v_2)}$ drawn from MNIST (left half) and USPS (right half). The unobserved samples are $x_i^{(v_1)}$ which is invisible to our model. The reconstructions and  imputations are the $\hat{x}_i^{(v_2)} = g(f(x_i^{(v_2)})$ and $\hat{x}_i^{(v_1)}$ in Algorithm~1 line 16 in Supplementary respectively. }
\label{Fig.6}
\vspace{-0.3cm}
\end{figure}

\begin{figure*}
\centering
\includegraphics[width=0.85\textwidth]{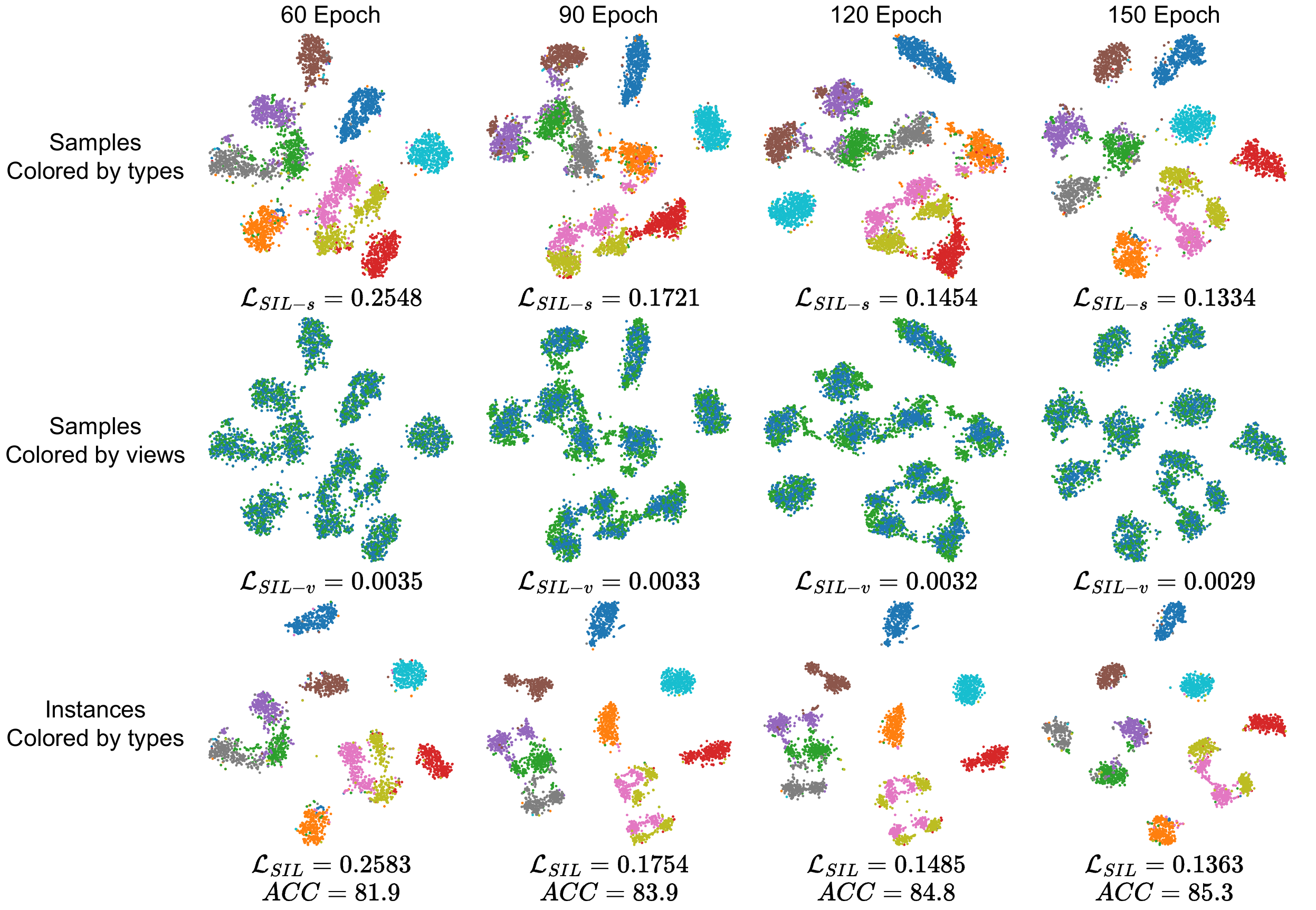}
\caption{\highlight{The t-SNE visualization of the clustering quality on NoisyMNIST with the unpaired rate of $100\%$.} The first two rows visualize the hidden representations of the samples that are colored according to their types (up) and views (middle) respectively. The last row (bottom) visualizes the hidden representations of the instances that are colored according to their types. } 
\label{Fig.8}
\vspace{-0.3cm}
\end{figure*}

\subsection{In-depth Explorations}
\label{Sec.4.3}

In this section, we conduct in-depth explorations to experimentally demonstrate the effectiveness of our SMILE and provide support for Theorem~\ref{Theorem.3}, Theorem~\ref{Theorem.2}, and Theorem~\ref{Theorem.4}.

\textbf{Tackling the problem of fully incomplete information via semantic invariance learning.} We first demonstrate the effectiveness of semantic invariance learning $I(C; X|V)$ in addressing correspondence incompleteness and instance incompleteness, which are theoretically proven in Theorem~\ref{Theorem.3} and Theorem~\ref{Theorem.2}, respectively. 


For correspondence incompleteness, we visualize the similarity matrices in Fig.~\ref{Fig.4} to help understand the performance of our realignment approach. In the figure, CAR~\cite{yang2022robust} is adopted to evaluate the alignment rate at the category level, which is defined as follows:
\begin{equation}
    CAR = \frac{1}{N} \sum_i \varsigma(T(x_i^{(v_1)}), T(\hat{x}_i^{(v_2)})),
\end{equation}
where $\varsigma$ is the Dirichlet function and $\hat{x}_i^{(v_2)}$ represents the realigned cross-view counterpart of $x_i^{(v_1)}$. The figure shows that CAR increases as semantic invariance learning progresses (as $\mathcal{L}_{SIL}$ decreases), demonstrating that SIL facilitates the realignment of the correspondence-incomplete data. 

For instance incompleteness, we evaluate the impact of semantic invariance learning on the imputation performance on Caltech by using the Normalized Root Mean Square Error (NRMSE)~\cite{hotelling1992relations}, which evaluates the imputation error of the unobserved samples. As shown in Fig.~\ref{Fig.5}, both NRMSE and $\mathcal{L}_{SIL}$ decrease as the value of $\lambda_{SIL}$ increases. This trend suggests that the imputation error is minimized by emphasizing $\mathcal{L}_{SIL}$ (\textit{i.e.}, increasing $\lambda_{SIL}$). Overall, the figure suggests that semantic invariance learning helps compensate for instance incompleteness. 

In addition to the quantitative evaluation, we visualize the imputed samples in Fig.~\ref{Fig.6}. The figure shows that the imputed samples (last row) belong to the same category as the missing samples, even if the categories are not explicitly known to our model. Furthermore, the imputed samples are quite similar to the unobserved ones, despite the distinct styles across different views. We attribute this to our multi-branch design in Equation~(\ref{Eq.3.3.1.4}), which enables our model to learn view-specific styles independently. In brief, this figure confirms the effectiveness of our SMILE in compensating for instance incompleteness, \textit{i.e.} the ability to impute missing samples.

\textbf{Boosting clustering quality through semantic invariance learning.} 
Next, we verify experimentally that the semantic invariance learning  $I(C; X|V)$ bounds the lowest achievable clustering error rate, as proved in Theorem~\ref{Theorem.4}. To demonstrate this, we visualize the clustering quality in Fig.~\ref{Fig.8}. This figure illustrates that as $\mathcal{L}_{SIL-s}$ decreases (first row), our SMILE learns more compact and balanced clusters. Additionally, as $\mathcal{L}_{SIL-v}$ decreases (second row), our method learns more semantic-invariant clusters. By combining these benefits through $\mathcal{L}_{SIL}$, SMILE effectively mitigates cross-view discrepancies while enhancing semantic discrimination across different categories (third row). This confirms the ability of our SMILE to improve clustering quality by leveraging semantic invariance learning.

\subsection{Ablations}
\label{Sec.4.4}
\begin{table}
  \caption{Ablation study of our $\mathcal{L}_{DAR}$, and $\mathcal{L}_{SIL}=\mathcal{L}_{SIL-s} + \gamma \mathcal{L}_{SIL-v}$ on NoisyMNIST. The $\mathcal{L}_{Rec}$ is defined in Equation~(\ref{Eq.4.4.1}).}
  \label{Tab.2}
  \centering
  \setlength{\tabcolsep}{3pt}
\resizebox{0.9\linewidth}{!}{ 
\begin{tabular}{c|cccc|ccc}
   \toprule
        Data Type& $\mathcal{L}_{Rec}$ &$\mathcal{L}_{DAR} $ & $ \mathcal{L}_{SIL-s}$ & $  \mathcal{L}_{SIL-v} $ & ACC & NMI & ARI   \\ \midrule
        \multirow{5}{*}{\highlight{\shortstack{$100\%$ Unaligned\\($\zeta=100\%$)}}} & $\checkmark$ & & & &50.1 & 47.7 & 32.3  \\ 
        &$\checkmark$& & $\checkmark$& &  52.7 & 54.5 & 39.1  \\ 
        &$\checkmark$& & &$\checkmark$&   74.5 & 66.1 & 58.4  \\ 
        &$\checkmark$& & $\checkmark$& $\checkmark$& \underline{78.2} & \underline{76.3} & \underline{69.0}  \\ 
        & & $\checkmark$ & $\checkmark$ & $\checkmark$ &  \textbf{82.7} & \textbf{79.5} & \textbf{74.2}  \\ \midrule  
        \multirow{5}{*}{\highlight{\shortstack{$100\%$ Missing\\($\eta=100\%$)}}} &$\checkmark$ & & & & 48.7 & 43.7 & 30.9  \\ 
        &$\checkmark$& & $\checkmark$& &  51.0 & 51.6 & 37.8  \\ 
        &$\checkmark$& & &$\checkmark$&   66.8 & 57.1 & 49.3  \\ 
        &$\checkmark$& & $\checkmark$& $\checkmark$& \underline{67.8} & \underline{63.5} & \underline{52.6}  \\ 
        & & $\checkmark$ & $\checkmark$ & $\checkmark$ &  \textbf{69.0} & \textbf{63.8} & \textbf{54.1}  \\ 
           \bottomrule
    \end{tabular}
    }
    \vspace{-0.3cm}
\end{table}

In this section, we present an ablation analysis to elucidate the mechanism of our SMILE. As shown in Table~\ref{Tab.2}, the performance of a standard auto-encoder alone (first row) is poor with the unaligned rate of $100\%$ and missing rate of $100\%$. However, when we introduce $\mathcal{L}_{SIL-v}=I(C; V)$ (third row), the performance is significantly boosted ($\ge 18\%$ for ACC). We conjecture that $\mathcal{L}_{SIL-v}$ helps alleviate the cross-view discrepancy, which is essential for learning consensus semantics for MvC. Moreover, the performance is further improved when combined with $\mathcal{L}_{SIL-s}=I(C; X)$ (fourth row), which enhances the semantic discrimination. Finally, by introducing the discrepancy-aware reconstruction term $\mathcal{L}_{DAR}$ in the fifth row, with the unaligned rate of $100\%$ and missing rate of $100\%$, we improve ACC by $4.5\%$ and $1.2\%$ respectively. This verifies the effectiveness of each component in SMILE.

\begin{figure} 
\centering
\includegraphics[width=1\linewidth]{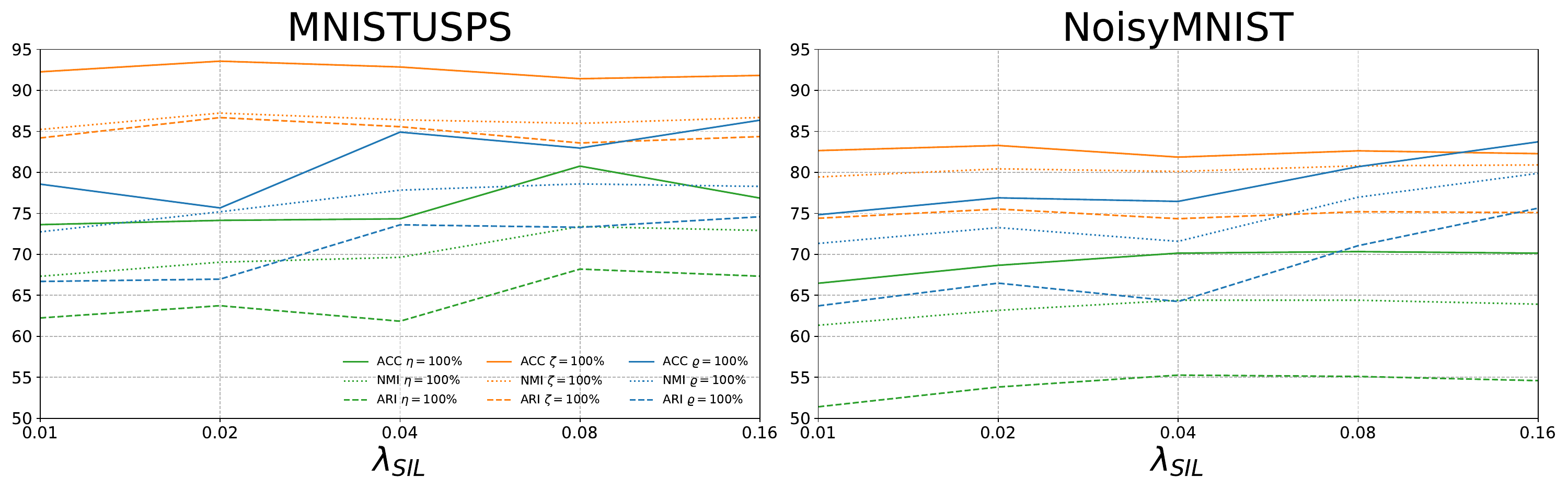}
\caption{Parameter analysis of $\lambda_{SIL}$ on \highlight{MNISTUSPS and NoisyMNIST with the unaligned rate ($\zeta$) of $100\%$, the missing rate ($\eta$) of $100\%$, and the unpaired rate ($\varrho$) of $100\%$.}}
\label{Fig.MNISTUSPSNoisyMNIST_lambdaSIL_MissingUnalignUnpair_ACCNMIARI}
\end{figure}

\begin{figure}
\centering
\includegraphics[width=1\linewidth]{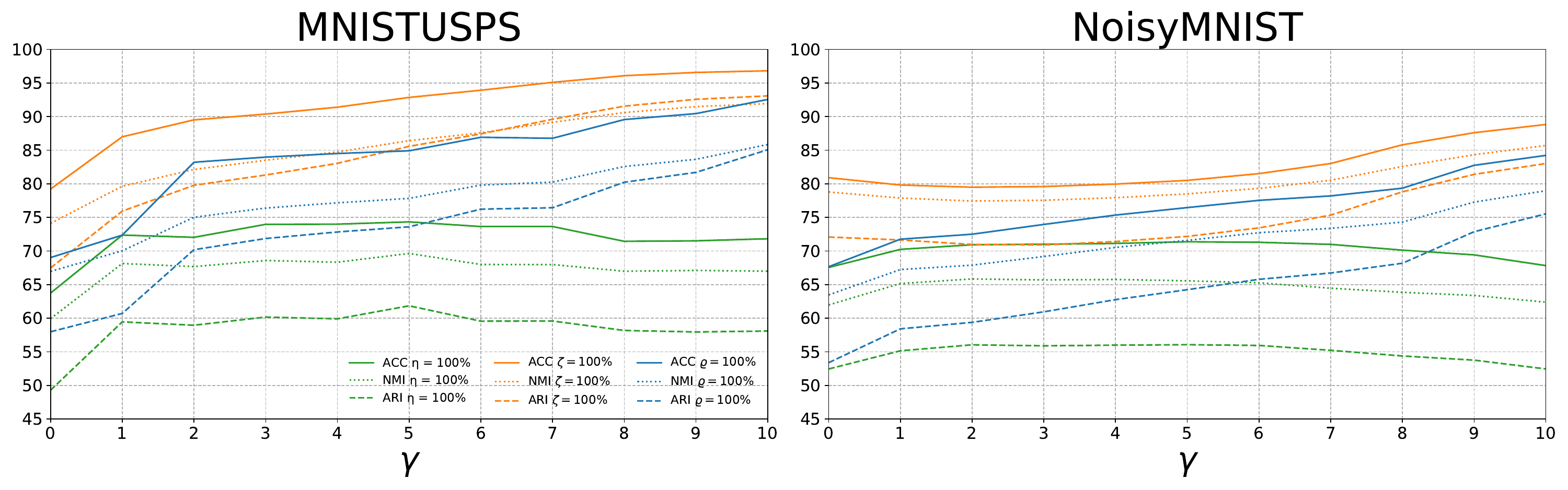}
\caption{Parameter analysis of $\gamma$ on \highlight{MNISTUSPS and NoisyMNIST with the unaligned rate ($\zeta$) of $100\%$, the missing rate ($\eta$) of $100\%$, and the unpaired rate ($\varrho$) of $100\%$.}}
\label{Fig.MNISTUSPSNoisyMNIST_gamma_MissingUnalignUnpair_ACCNMIARI}
\end{figure}

To investigate the influence of the parameters, \highlight{we conduct parameter analysis in Figs.~\ref{Fig.MNISTUSPSNoisyMNIST_lambdaSIL_MissingUnalignUnpair_ACCNMIARI} and \ref{Fig.MNISTUSPSNoisyMNIST_gamma_MissingUnalignUnpair_ACCNMIARI}. As shown in the figures, the SMILE performs stably against the hyper-parameters $\lambda_{SIL}$ and $\gamma$ under the three settings. Besides, one could observe that the performance remarkably drops when $\gamma = 0$, indicating the importance of semantic-invariance learning. }



\section{Conclusion}
In this paper, we addressed a challenging problem of multi-view clustering with fully incomplete information. To the best of our knowledge, this could be one of the first studies on the challenge. We propose a foundational theorem, Semantic Invariance, that enables us to alleviate the cross-view discrepancy based on their semantic distributions without the requirement for paired samples, thus learning consensus semantics. Building on this theorem, we proposed a unified semantic invariance learning framework for MvC with fully incomplete information. We showed, both theoretically and experimentally, that our framework could not only effectively compensate for incomplete information, but also facilitate MvC. Specifically, our SMILE achieved superior performance compared to 13 state-of-the-art baselines under various incomplete settings on five benchmarks. In the future, we would like to endow our method with the ability to handle more practical scenarios where incomplete information occurs unconsciously, and the incomplete instances/correspondences are unknown. This will allow us to apply our method to a wider range of real-world problems.


%





\ifCLASSOPTIONcaptionsoff
  \newpage
\fi



%
\bibliographystyle{IEEEtran}
\bibliography{reference}



%




\begin{IEEEbiography}[{\includegraphics[width=1in,height=1.25in,clip,keepaspectratio]{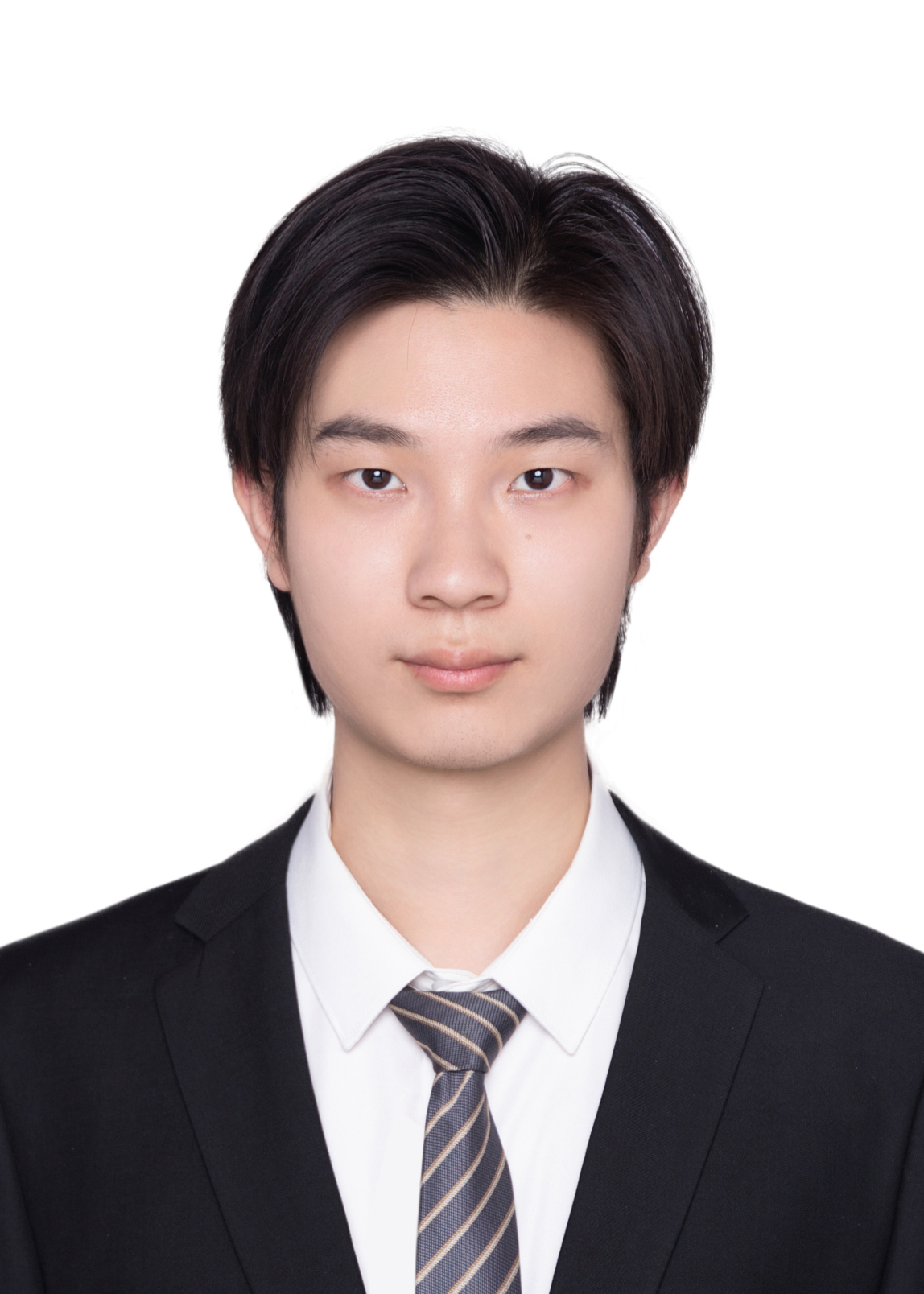}}]{Pengxin Zeng}
received the bachelor’s degree in computer science and technology from Sichuan University, Chengdu, China, in 2021, where he is currently pursuing the master’s degree in computer science with the College of Computer Science. His research interest includes 3D vision and multi-modality learning.
\end{IEEEbiography}
\begin{IEEEbiography}[{\includegraphics[width=1in,height=1.25in,clip,keepaspectratio]{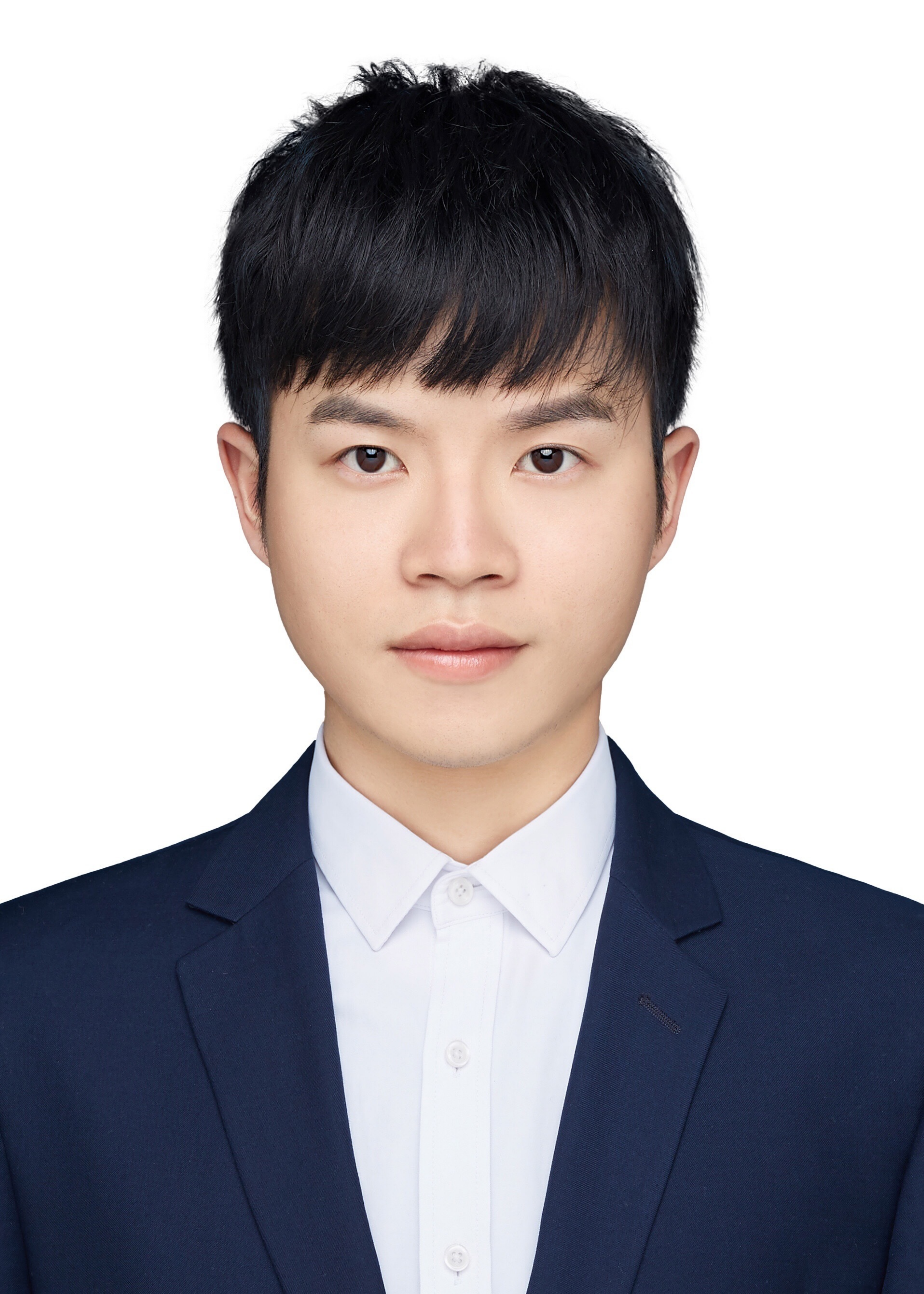}}]{Mouxing Yang}~received the bachelor’s degree in computer science and technology from Sichuan University, Chengdu, China, in 2020, where he is currently pursuing the Ph.D. degree in computer science with the College of Computer Science. His research interest includes multi-modal representation learning.
\end{IEEEbiography}
\begin{IEEEbiography}[{\includegraphics[width=1in,height=1.25in,clip,keepaspectratio]{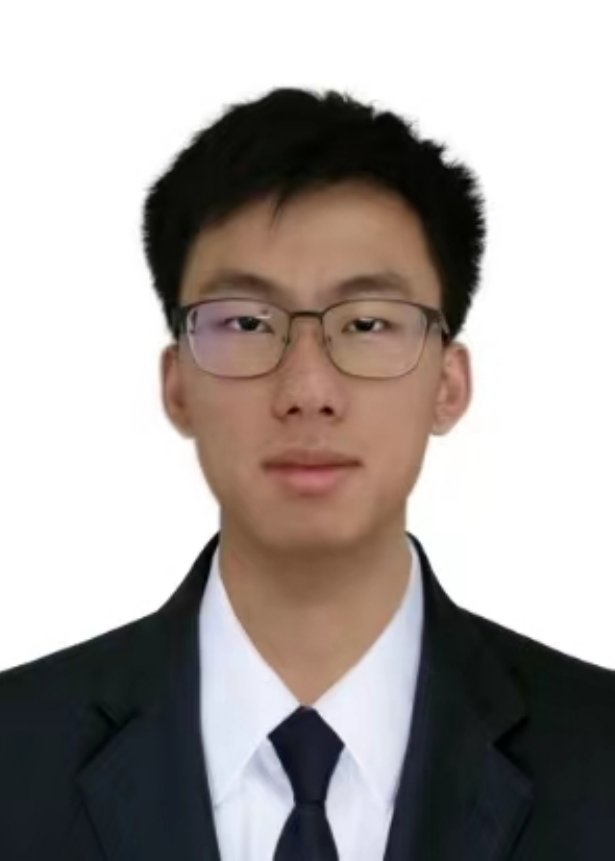}}]{Yiding Lu}
received the bachelor’s degree in computer science and technology from Sichuan University, Chengdu, China, in 2022, where he is currently pursuing the master’s degree in computer science with the College of Computer Science. His research interest includes multi-view learning and unsupervised learning.
\end{IEEEbiography}
\begin{IEEEbiography}[{\includegraphics[width=1in,height=1.25in,clip,keepaspectratio]{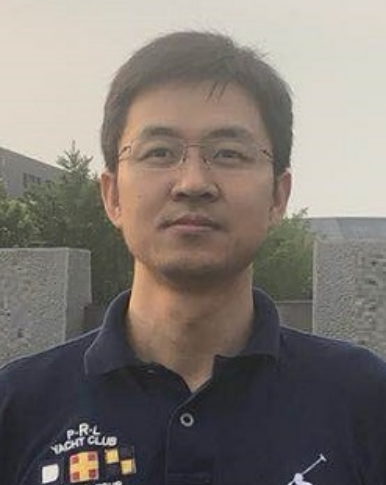}}]{Changqing Zhang}
received the B.S. and M.S. degrees from the College of Computer, Sichuan University, Chengdu, China, in 2005 and 2008, respectively, and the Ph.D. degree in Computer Science from Tianjin University, China, in 2016. He is an associate professor in the College of Intelligence and Computing, Tianjin University. He has been a postdoc research fellow in the Department of Radiology and BRIC, School of Medicine, University of North Carolina at Chapel Hill, NC, USA. His current research interests include machine learning, computer vision and medical image analysis.
\end{IEEEbiography}
\begin{IEEEbiography}[{\includegraphics[width=1in,height=1.25in,clip,keepaspectratio]{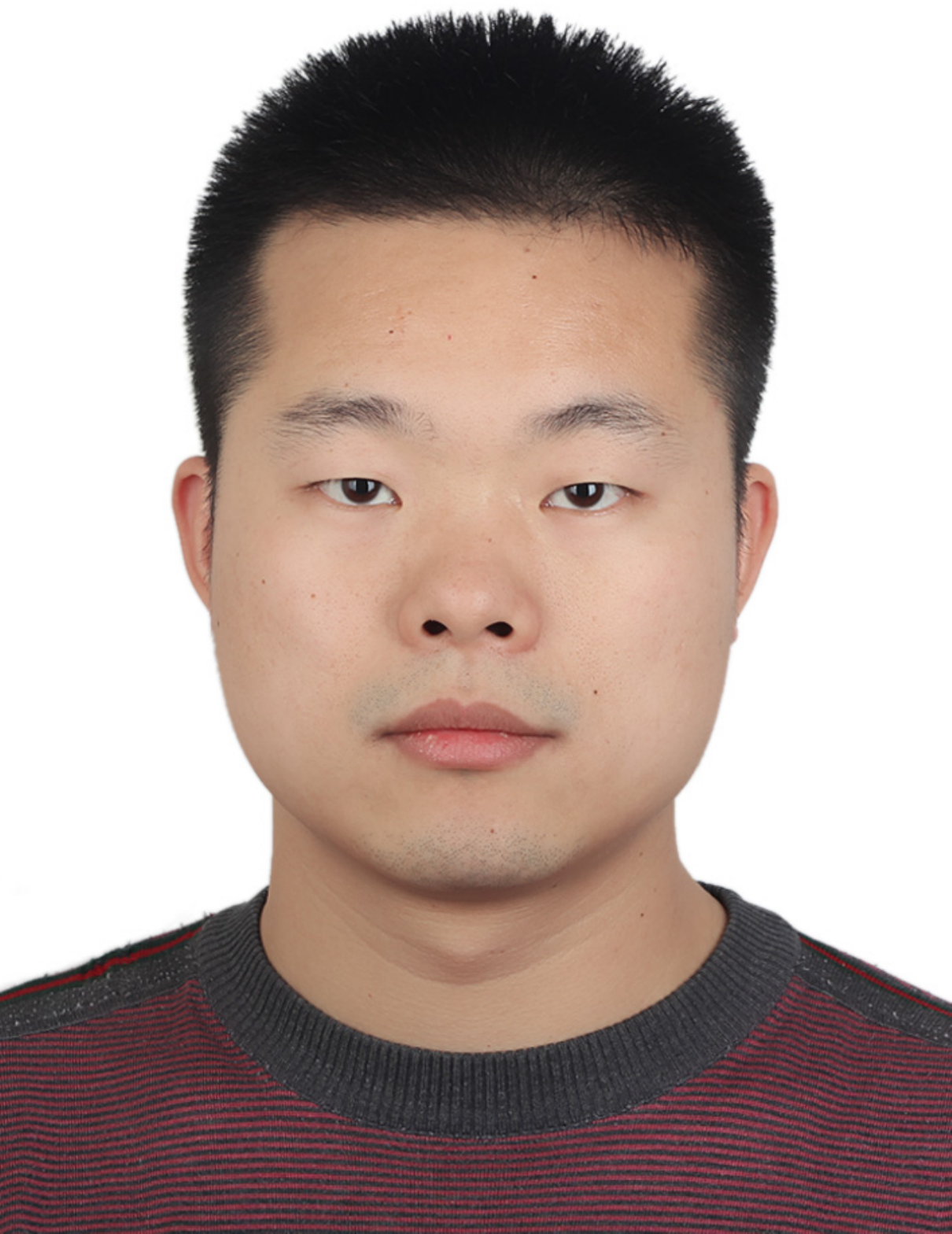}}]{Peng Hu}
received the Ph.D. degree in computer science and technology from Sichuan University, China, in 2019. He is currently an associate research professor at the College of Computer Science, Sichuan University. His research interests mainly focus on multi-view learning, cross-modal retrieval, and network compression. On these areas, he has authored more than 20 articles in the top-tier conferences and journals.
\end{IEEEbiography}
\begin{IEEEbiography}[{\includegraphics[width=1in,height=1.25in,clip,keepaspectratio]{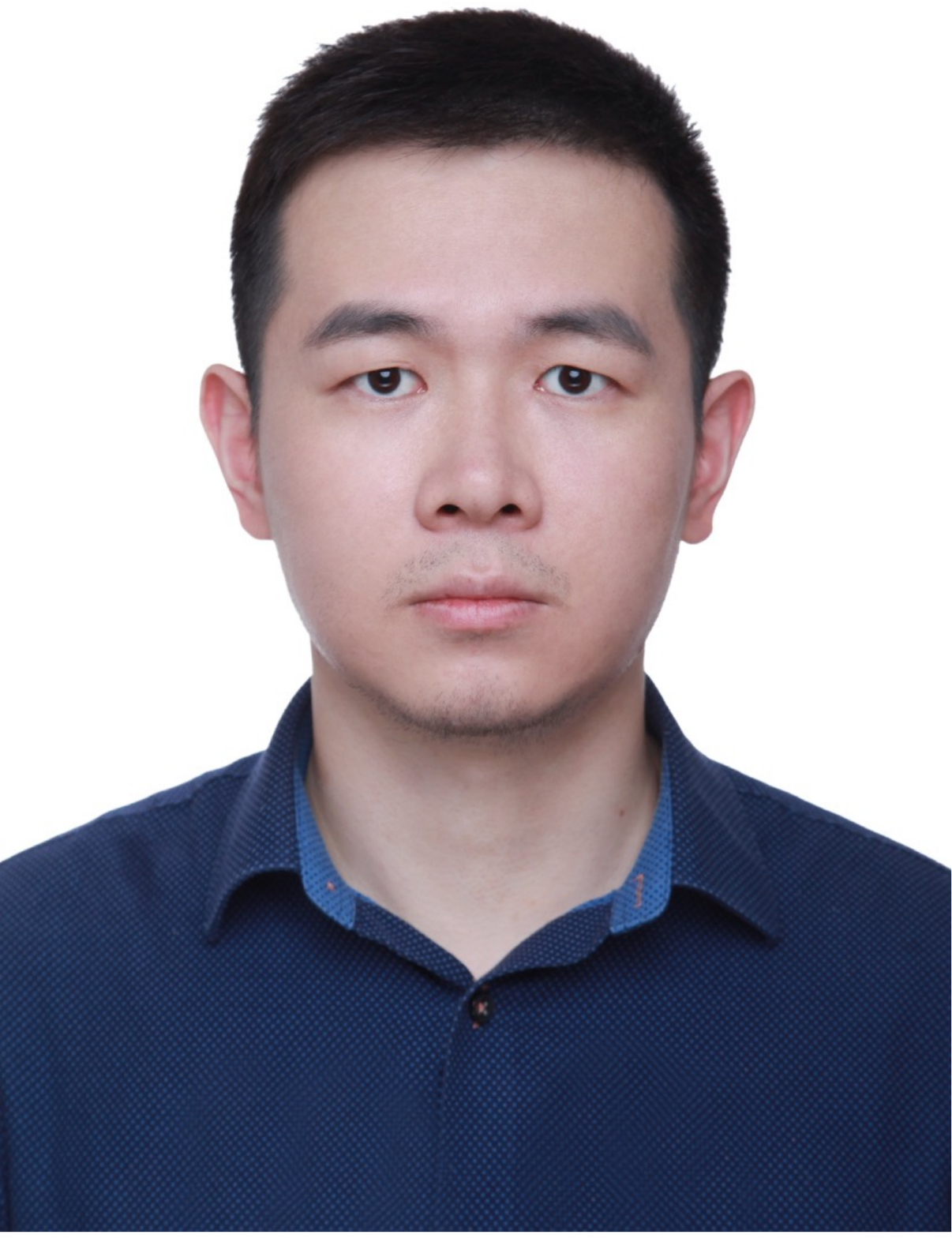}}]{Xi Peng}
is currently the Cheung Kong distinguished professor at the College of Computer Science, Sichuan University. His current interests mainly focus on machine learning, multi-media analysis, and AI4Science. In these areas, he has authored around 100 articles in Nature Communications, JMLR, TPAMI, ICML, NeurIPS, and so on. Dr. Peng has served as an Associate Editor for four journals such as “ IEEE Trans on SMC: Systems”, and a Guest Editor for four journals such as “IEEE Trans. on Neural Network and Learning Systems”.
\end{IEEEbiography}

\end{document}